\documentclass[12pt,letterpaper]{article}
\usepackage[a4paper, total={7in, 10in}]{geometry}

\usepackage{graphicx}
\usepackage{helvet}
\usepackage{authblk}
\usepackage{hyperref}
\usepackage{amsmath} 
\usepackage{amssymb} 
\usepackage{orcidlink} 
\usepackage{float}
\usepackage[linesnumbered,ruled]{algorithm2e}
\usepackage[super,comma,sort&compress]{natbib}
\usepackage[right]{lineno} \linenumbers
\usepackage{soul}

\makeatletter
\renewcommand{\maketitle}{\bgroup\setlength{\parindent}{0pt}
\begin{flushleft}
  {\Huge \textbf{\@title}}
  
  \@author
\end{flushleft}\egroup}
\makeatother

\title{Insect-Wing Structured Microfluidic System for Reservoir Computing}
\date{}

\begin{document}
\nolinenumbers

\author[1]{Jacob Clouse}
\author[2]{Thomas Ramsey}
\author[1] {Samitha Somathilaka}
\author[1] {Nicholas Kleinsasser}
\author[2] {Sangjin Ryu}
\author[1] {Sasitharan Balasubramaniam}

\affil[1]{School of Computing, University of Nebraska-Lincoln, Lincoln, Nebraska, USA}
\affil[2]{Department of Mechanical and Materials Engineering, University of Nebraska-Lincoln, Lincoln, Nebraska, USA}

\affil[*]{Correspondence: jclouse2@huskers.unl.edu, tramsey3@huskers.unl.edu, ssomathilaka2@unl.edu, nkleinsasser2@huskers.unl.edu, sryu2@unl.edu, sasi@unl.edu}

\maketitle

\section{The Bigger Picture}

Reservoir computing is a highly efficient machine learning approach that leverages fixed internal weights to process and apply non-linear interactions to data. Meanwhile, microfluidic computing drawn from biological forms has introduced new novel physical mediums for computation. Both paradigms offer alternative and potentially more efficient computing methods. By combining them, we propose a hybrid system that merges the strengths of each to create a unique, efficient device tailored for artificial intelligence (AI) applications. Specifically, we present a bio-inspired microfluidic chip that functions as a reservoir to perform pattern recognition tasks. This end-to-end system demonstrates how patterns, represented as red, green, and blue (RGB) dyes, are transformed and classified through a sequence involving fluid injection systems, a microfluidic chip, a camera, and a final classification step performed in software. This type of device opens up opportunities for alternative forms of computation that could be run in parallel and benefit traditional silicon systems.

\section{SUMMARY}

As the demand for more efficient and adaptive computing grows, nature-inspired architectures offer promising alternatives to conventional electronic designs. Microfluidic platforms, drawing on biological forms and fluid dynamics, present a compelling foundation for low-power, high-resilience computing in environments where electronics are unsuitable. This study explores a hybrid reservoir computing system based on a dragonfly-wing inspired microfluidic chip, which encodes temporal input patterns as fluid interactions within the micro channel network.

The system operates with three dye-based inlet channels and three camera-monitored detection areas, transforming discrete spatial patterns into dynamic color output signals. These reservoir output signals are then modified and passed to a simple and trainable readout layer for pattern classification. Using a combination of raw reservoir outputs and synthetically generated outputs, we evaluated system performance, system clarity, and data efficiency. The results demonstrate consistent classification accuracies up to $91\%$, even with coarse resolution and limited training data, highlighting the viability of the microfluidic reservoir computing.

\section*{KEYWORDS}

Reservoir Computing, Microfluidics, Bio-Inspired Engineering, Classification. 

\section{INTRODUCTION}

Reservoir computing (RC)\cite{gauthier_next_2021, uzun_molecular_2025, bai_design_2023, nakajima_physical_2022, raab_brownian_2022, matsumura_real-time_2025} is an efficient and flexible computational paradigm designed to process sequential and temporal data. It originates from recurrent neural network (RNN) models \cite{mienye_recurrent_2024, hibat-allah_recurrent_2020, maass_real-time_2002}, but offers a significant simplification. The internal network, known as the reservoir, is made up of a fixed, high-dimensional, nonlinear dynamic system that projects input signals into a rich feature space. This transformation of the data in the reservoir is represented as a black box\cite{loyola-gonzalez_black-box_2019, hassija_interpreting_2024}, modifying the data in such a way that the features are extracted and made more identifiable for a simple machine learning model. Only the readout layer is trained, typically through simple linear regression, which drastically reduces the computational cost of training compared to traditional RNNs. This architecture retains the temporal memory and dynamic behavior needed for complex tasks such as time series forecasting \cite{xue_promptcast_2024, koster_data-informed_2023, canaday_rapid_2018} and pattern recognition \cite{zhang_towards_2020, kopets_simulation_2022, tsakalos_protein_2022}, while eliminating many of the challenges associated with training recurrent connections.

RC has evolved beyond its original software-based implementations and is now realized on a broad spectrum of physical substrates, including photonic systems\cite{garcia-beni_scalable_2023, chembo_machine_2020}, spintronic devices\cite{misba_spintronic_2023, furuta_macromagnetic_2018}, memristive circuits \cite{zhong_memristor-based_2022, chen_full-analog_2024, fu_enabling_2024, sun_-sensor_2021, zhong_dynamic_2021}, gas sensors\cite{jiang_-sensor_2025}, and even biological platforms \cite{cai_brain_2023, angerbauer_novel_2024, angerbauer_molecular_2024, nikolic_computational_2023, sumi_biological_2023}. Each of these models exploits the intrinsic dynamics of the medium to function as a reservoir. Such physical realizations offer the potential for ultra-fast, energy-efficient computation. The simplicity and efficiency of RC continues to position it as a powerful tool for next-generation computing systems that require real-time processing of complex and time-varying signals. In this paper, we introduce a novel RC substrate that utilizes a \emph{polydimethylsiloxane} (PDMS) microfluidic chip inspired by the vein network of a dragonfly wing\cite{ryu_insect_2025}.

In recent years, microfluidics has been geared toward the medical and biological industries with advances such as organ-on-chip \cite{leung_guide_2022, bhatia_microfluidic_2014} and lab-on-chip \cite{garcia-hernandez_optical_2023, temiz_lab---chip_2015, gomez_communicating_2025}. An attractive aspect of microfluidic systems is the ability to replicate veins and channels of an organism on a chip \cite{miali_leaf-inspired_2019, ansarizadeh_microfluidic_2025}. Although the vast majority of applications are in the biological or chemical domains, developments have been made using microfluidic systems for computing and communication\cite{kuscu_modeling_2016, kuscu_fabrication_2021, zhang_ion_2025}. For example, boolean logic gates have been proposed and created using microfluidic chips\cite{singh_logic_2023, elatab_pressuredriven_2020, azizbeigi_microfluidic-based_2021}, inspiring us to create a system using a microfluidic chip for RC.

Our proposed system is illustrated in Figure~\ref{fig:Figure1}(A) and is designed to classify input patterns into eight categories. When entering the microfluidic RC chip, each input to the system is encoded from a pattern into a red, green, or blue (RGB) dye, which is achieved by using fluid injection systems consisting of syringes, tubing, and syringe pumps. The chip we use is pre-filled with clear water so that the different dyes can flow freely and propagate throughout the wing vein network structure\cite{ryu_insect_2025}. Next, we utilize the unique network structure inspired by the wing vein system of dragonfly (\emph{Common Green Darner}), to mix different RGB dyes, which creates nonlinear interactions between inputs. A camera positioned over the reservoir is used to record three specific detection areas, where python code is then utilized to detect dye, breakdown, and produce nine unique color output signals from the reservoir. The output signal of each color from the detection areas is then quantized before it is sent to the readout layer as a spatio-dataset, as shown in Figure~\ref{fig:Figure1}(A). This trainable readout layer is located on a conventional computer and will be used to classify the dataset into eight different patterns. This combination results in a novel hybrid RC system with Figure~\ref{fig:Figure1}(B) showing the end-to-end setup. Figure~\ref{fig:Figure1}(B) also shows a close-up view of the microfluidic reservoir chip mimicking the dragonfly wing vein structure. In this paper, we will demonstrate how our hybrid RC system is used to detect $3 \times 5$ patterns as a proof-of-concept for hybrid computing using insect-inspired microfluidics.

\section{RESULTS}
\label{sec:Results}

\begin{figure}[H]
    \centering
    \includegraphics[width=1\textwidth]{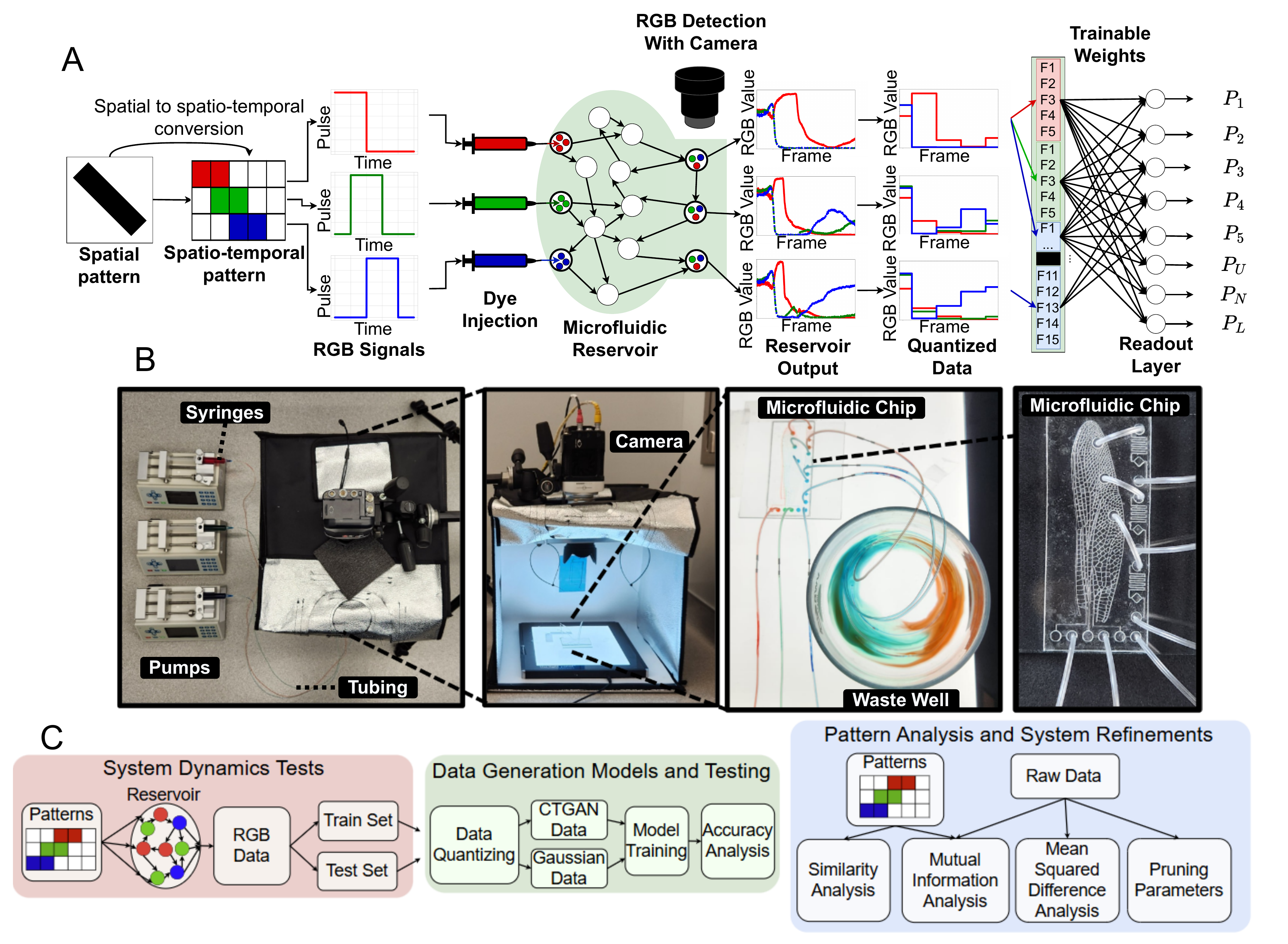}
    \caption{
    (A) A schematic overview of the system workflow, illustrating the transformation of spatial patterns into spatio-temporal input signals, which are delivered to a microfluidic chip. The resulting dynamic outputs are captured by a camera, processed into features through area detection and quantization, and finally classified in a trainable readout layer. (B) A complete laboratory setup. The first panel shows a top-down view with three input syringe pumps, the dragonfly wing chip, and an overhead camera. The second panel presents a front view of the setup, highlighting the downward-facing camera and a light box used for illumination. The third panel depicts the microfluidic chip during fluid flow, while the final panel shows the same device with tubing attached but no fluid present.
    (C) Experimental workflow and manuscript structure. The study begins with characterization of the system dynamics, followed by data generation and model testing, and concludes with a pattern analysis phase to evaluate classification behavior.}
    \label{fig:Figure1}
\end{figure}

The structure of the following sections is illustrated in Figure~\ref{fig:Figure1}(C), beginning with foundational experiments and advancing through successive stages of system refinement. We start our investigation by performing system dynamics tests on the microfluidic reservoir utilizing a series of input patterns to characterize its baseline behavior in Section~\ref{sec:Chip Dynamics and System Behavior}. This initial stage focuses on how different colored dyes propagate through the network of channels, providing information on temporal retention, mixing characteristics, and color sensitivity. We end this section with data split into train and test sets to be used in subsequent sections. Building on these observations, we transition from physical experimentation to computational processing in Section~\ref{sec:Pre-Processing Data and Readout Layer} which covers data generation models, model training, and an analysis of the accuracy of the system. In this phase, we apply preprocessing steps to the collected data, including spatio-temporal data conversion applied through quantization and generation of new reservoir output signals via synthetic data generation. Subsequently, a readout layer is developed and trained to classify the resulting spatio-dataset.

\begin{figure}[H]
    \centering
    \includegraphics[width=1\textwidth]{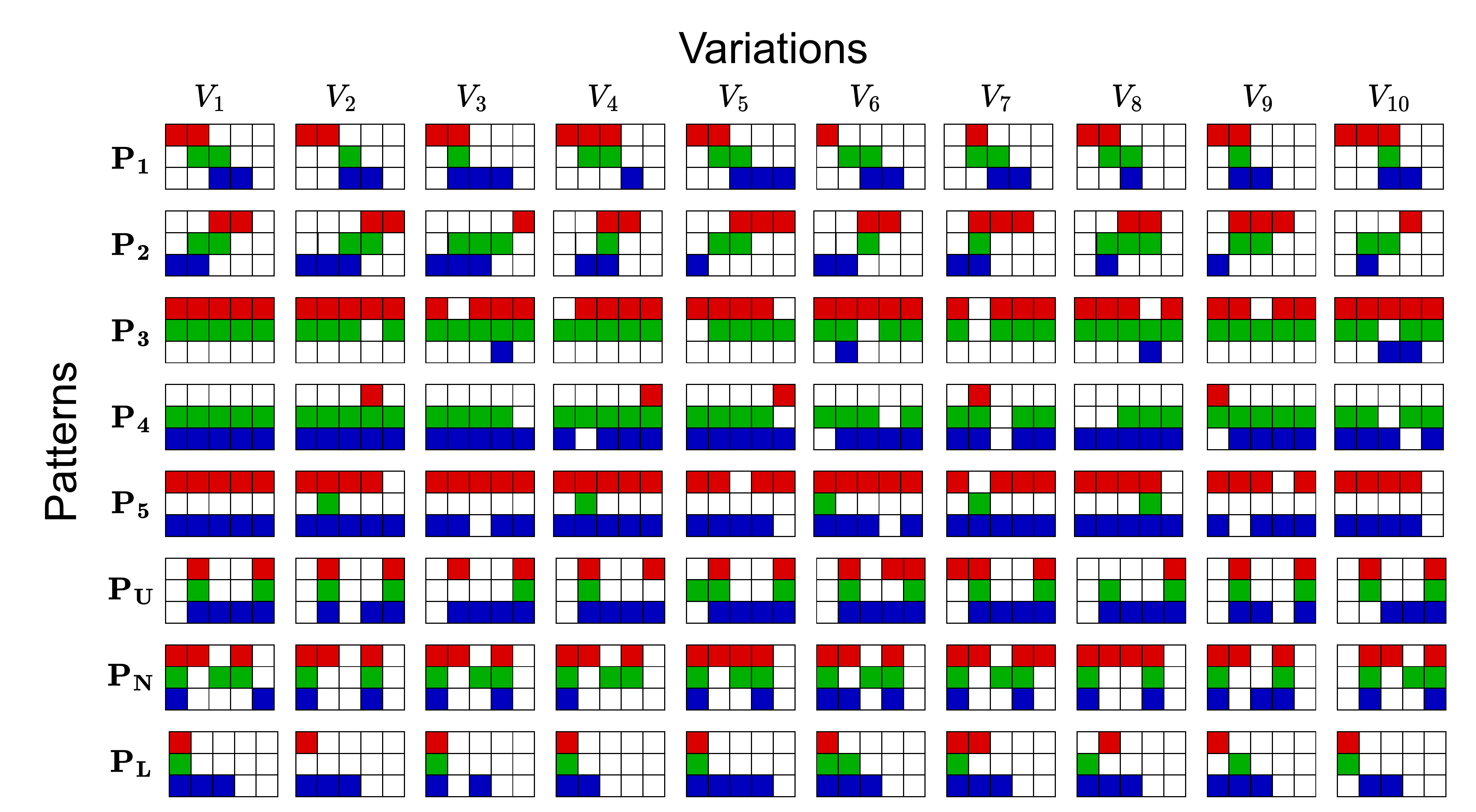}
    \caption{
    A composite display of all input samples presented to the reservoir for classification. The input data includes eight distinct pattern types: five shape patterns ($\bf{P_1}$, $\bf{P_2}$, $\bf{P_3}$, $\bf{P_4}$, $\bf{P_5}$), and three letter patterns ($\bf{P_U}$, $\bf{P_N}$, $\bf{P_L}$) representing letters ($'U'$, $'N'$, $'L'$). Each pattern type has ten variations with minor augmentations, and $P_{(i,V_j)}$ denotes $j^{th}$ variant of pattern $\mathbf{P_i}$.
    }
    \label{fig:Figure2}
\end{figure}

Finally, we conduct a series of pattern analysis refinements to improve model accuracy and system efficiency. These efforts are presented in Section~\ref{sec:Dimensionality Reduction of Readout Layer}, where we analyze the mutual information (MI) between patterns input to the reservoir and output signals from the reservoir. This provides us with insight on how input patterns affect information flow. We then evaluate the contribution of individual reservoir output areas to analyze priority regions of the chip and their role in classifying patterns. Next, we evaluate the effect of white balancing in order to determine if modifying the raw output signals of the reservoir can show the reliability of the system against different lighting conditions. We conclude Section~\ref{sec:Dimensionality Reduction of Readout Layer} with a targeted input pattern analysis to better understand the remaining sources of classification error and the limitations imposed by system resolution. The different types of patterns, as well as their variations, are shown in Figure~\ref{fig:Figure2}. 

\begin{figure}[H]
    \centering
    \includegraphics[width=1\linewidth]{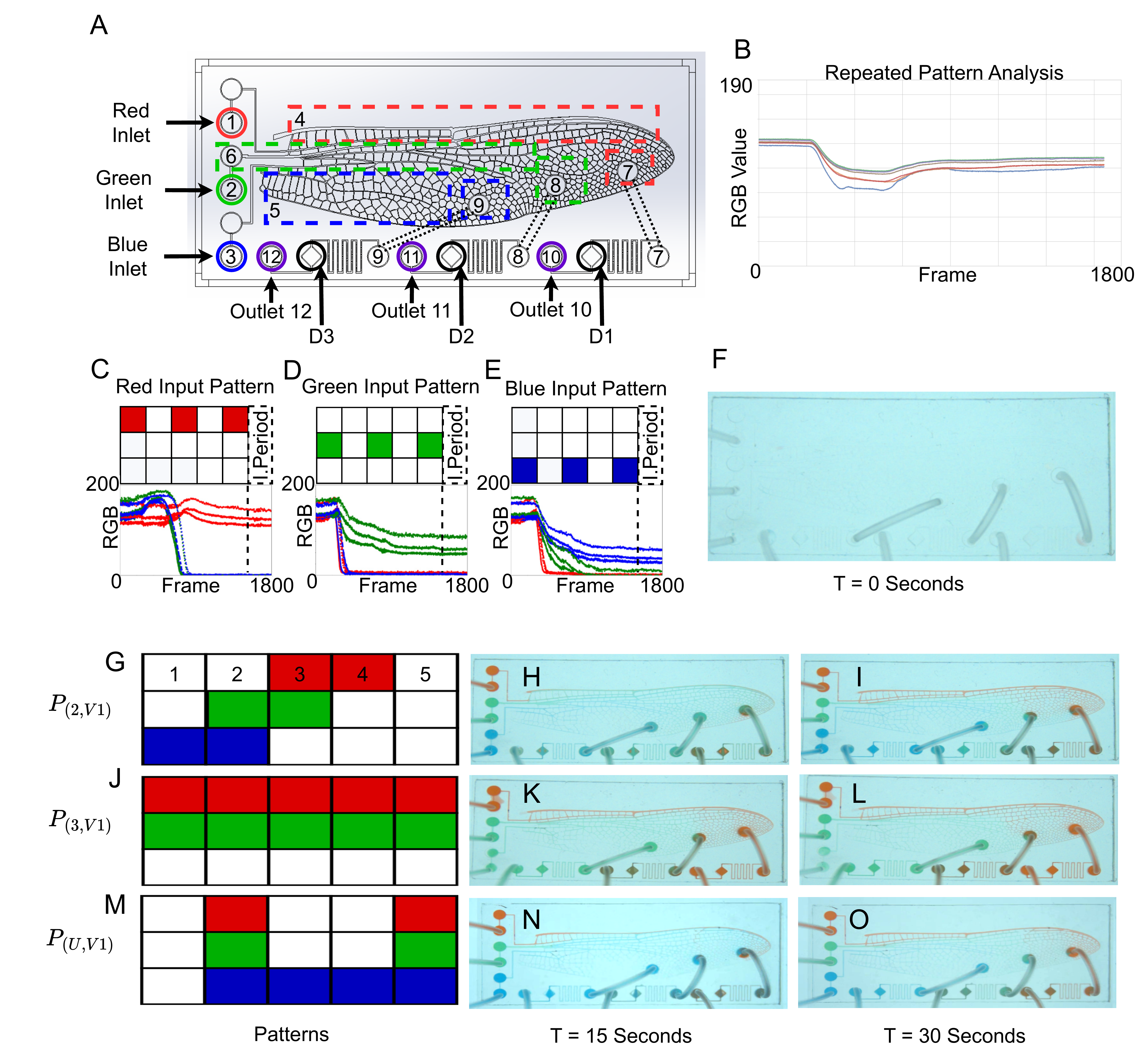}
    \caption{
    (A) An image of the microfluidic chip showing important regions of analysis for the dye behavior in the system including inlets, outlets, and propagation areas (See the final panel in Figure~\ref{fig:Figure1}(B)).
    (B) Trial done to show system reliability by repeating one pattern five times.
    (C,D,E) Evaluating dye retention in the system by sending pulses of red, green, and blue dye. Each pattern covers $1500$ frames ($30$ seconds) of each graph with the last $300$ frames (five seconds) being an idling period where no pattern is injected and the system remains stationary.
    (F - Q) Three trials done with patterns $\bf{P_2}$, $\bf{P_3}$, and $\bf{P_U}$ to show dye mixing in the microfluidic chip over $30$ seconds.
    }
    \label{fig:Figure3}
\end{figure}

\subsection{Chip Dynamics and System Behavior}
\label{sec:Chip Dynamics and System Behavior}

\subsubsection{Observation Areas and System Behavior With Repeated Signals}
\label{sec:Observation Areas and System Behavior With Repeated Signals}
To evaluate the feasibility of using a microfluidic reservoir for pattern-based computation, we begin by characterizing the raw behavior of the system under controlled input conditions. This phase serves as the foundation for understanding how different fluid signals propagate through the network and influence downstream output. Using three inlet ports, each carrying red, green, or blue dye, we inject binary encoded patterns shown in Figure~\ref{fig:Figure2} into the reservoir and observe how the system responds over time. As shown in Figure~\ref{fig:Figure3}(A), the inlet ports are located along the left side of the device, denoted by $1$ for the red port, $2$ for the green port, and $3$ for the blue port, while the detection areas are represented as three diamond-shaped areas located at the bottom of the wing denoted by $D1$, $D2$, $D3$. The three boxed areas on the chip represent the primary flow paths of different dyes, labeled $4$–$9$, with areas $7$–$9$ connected to separate detection regions via tubes and a mixing channel, as indicated by two dotted lines in Figure~\ref{fig:Figure3}(A). The outlet ports for the system are $10$, $11$, and $12$. This direct mapping enables a clear interpretation of dye behavior by labeling each inlet port, specific detection area, and outlet port, allowing for straightforward analysis of signal propagation, retention, and mixing behavior within the reservoir.

When only one dye is injected, it can spread freely throughout the chip, and when two dyes are injected, they also disperse more broadly across the chip. When all dyes are injected, output area $7$ primarily contains red, area $8$ exhibits a mixture of all three dyes, and area $9$ is dominated by blue. This behavior is a product of the way dyes are injected with larger channels carrying the red dye to output area $7$ through area $4$ and the close distance between the blue inlet port $3$ and area $9$ leading to a higher concentration of blue dye at this location. However, because of fluid mixing and variations in dye injection times, all output areas contain measurable concentrations of multiple dyes during various stages of pattern injection.

As patterns are sent through the system, a video is taken of the entire chip at $60$ frames per second ($fps$) for $30$ seconds. Each vertical cell in the patterns takes up five seconds of the video ($300$ frames). The last five seconds there is no pattern injection occurring. To quantify the output of the microfluidic chip, nine RGB values are extracted using a custom Python script making use of the OpenCV (CV2) library\cite{opencv_library}. These values are taken from three previously defined detection areas, which remained fixed throughout the study to serve as primary data sources. These nine signals represent three spatial regions that are then quantized into discrete intervals to enable structured and simplified input for downstream classification. This process effectively performs a spatial expansion of the inputs that are sent through the reservoir.

To evaluate the reliability of the system, identical input signals were injected five times and the reservoir output signals were observed and shown in Figure~\ref{fig:Figure3}(B). The high consistency across these repetitions demonstrates the computing reliability of the system, where identical inputs reliably produce identical outputs. Next, reservoir output signals from individual, fluctuating dyes being sent through the system are shown in Figure~\ref{fig:Figure3}(C–E). These tests reveal a key characteristic of the microfluidic reservoir, which is its tendency to retain elevated output signal intervals even after the input has been terminated. This behavior results from the absence of continuous flow. Once a dye is introduced, it remains stationary within the channel network unless it is displaced. As long as no new injection occurs, the reservoir output signal can persist indefinitely, effectively locking in the spatial distribution of the dye. This persistence suggests that the reservoir has a form of short-term memory similar to existing reservoir systems \cite{so_short-term_2023, liao_short-term_2023}, that is driven by fluid retention and the static nature of the medium.

Among the three color channels shown in Figure~\ref{fig:Figure3}(C–E), red and green reservoir output signals are consistently detected with high fidelity, while blue reservoir output signals appear weaker or mixed with green. This is likely a result of optical limitations in the camera or spectral overlap during image processing. Despite this limitation, each pattern produces a unique and distinguishable RGB signature that can later be used for classification.

Another important feature of the system is the contrast between “\emph{no dye}” and “\emph{full dye}” conditions. In the absence of dye, the clear fluid registers as white, producing high RGB values across all channels as shown in Figure~\ref{fig:Figure3}(C–E). Once a specific dye is introduced, the intensity in the dyes that are not injected decreases. This difference in response allows the system to distinguish minimal pattern variations, enabling downstream models to effectively separate inputs despite the low spatial resolution. We also observe that red dye takes two injection slots to reach detection areas compared to red and green dye being observed after the first injection has occurred.

\subsubsection{Observing Dye Movement With Input Patterns}
\label{sec:System Observation With Patterns}

To visualize the complete temporal progression of dye movement, we present the state of the system at three distinct time points for three different patterns, as shown in Figure~\ref{fig:Figure3}(F–O). At $T = 0$ seconds (visualized in Figure~\ref{fig:Figure3}(F)), the chip is filled only with water, representing a baseline condition without any input signals to the reservoir. This stage appears identical across all patterns. By $T = 15$ seconds, the first three grids of the encoded pattern have been injected. The first pattern we examine is shown in Figure~\ref{fig:Figure3}(G), displaying variant $P_{(2,V_1)}$, with two pulses of each dye. At $T = 15$ in Figure~\ref{fig:Figure3}(H), all three dyes are present: blue is concentrated in area $5$ and output area $9$, red appears faintly in area $4$ and output area $7$, and green is distributed across area $6$ and output areas $7$, $8$, and $9$, with the highest concentration in output area $8$. Next, at $T = 30$, Figure~\ref{fig:Figure3}(I) represents a balanced RGB pattern, with two time steps for each color present in the system. Red is found in area $4$ and output area $7$, blue remains in area $5$ and output area $9$, and green appears in area $6$, with an additional presence in output areas $8$ and $9$. All three dyes overlap in output area $8$. In the case of $\bf{P_2}$, when one dye is injected, while the other two are not, that dye forces out some of the previously injected dyes. This behavior can be compared in the difference between Figure~\ref{fig:Figure3}(H) and (I), and is observed whenever one or two dyes are injected while another is not, affecting the channels closest to the dye currently being injected.

Moving on to variant $P_{(3,V_1)}$ in Figure~\ref{fig:Figure3}(J), there are continuous reservoir input signals of red and green that are sent throughout the pattern. To visualize its behavior, in Figure~\ref{fig:Figure3}(K), red appears in area $4$ and output areas $7$ and $8$, while green is present in areas $5$ and $6$ as well as output areas $8$ and $9$. In this case, no blue dye is detected. When only two colors are present, they can reach more areas of the chip. In the case of $\bf{P_3}$, since the blue dye is not used, the green dye instead fills areas $5$ and $9$. The red dye also expands into output area $8$ more since the green dye flows through different channels. At $T = 30$ seconds in Figure~\ref{fig:Figure3}(L), there is minimal change compared to the previous frame. This occurs because $\bf{P_3}$ includes steady reservoir input signals of red dye and green dye with no new dye being introduced between time steps.

Variant $P_{(U,V_1)}$ is displayed in Figure~\ref{fig:Figure3}(M), representing the letter $'U'$. At $T = 15$ in Figure~\ref{fig:Figure3}(N), blue dye is more dominant, appearing strongly in area $5$ and output area $9$ with some presence in output area $8$. This is because more blue dye is injected into the system compared to red and green dyes. Once the red and green dyes stop flowing, the blue dye forces out a lot of the green dye in area $6$. The red dye is mainly localized faintly in area $4$ and output area $7$, and the green dye is faintly present in areas $6$ and $4$. At $T = 30$ seconds, the full pattern has propagated through the chip. This is shown in Figure~\ref{fig:Figure3}(O), where blue dye is less prevalent than at $T = 15$ seconds, while the red dye is more prevalent in area $4$ and the output area $7$. When the red dye is turned on in the last time step, it forces the green dye out of area $4$. The green dye is present in area $6$ and output area $8$ since it is also being injected. The takeaway from variant $P_{(U,V_1)}$ is that blue dye dominates most areas due to its greater representation in the encoded reservoir input signals in the middle time steps, but is forced out again once the red and green dye pumps are turned on. These three patterns show some unique behaviors in the system, with variant $P_{(2,V_1)}$ showing even changes in the system over time since two time steps of each color are injected, variant $P_{(3,V_1)}$ showing two continuous reservoir input signals over all time steps with red and green dyes in the pattern but no blue dye present, and variant $P_{(U,V_1)}$ showing blue dye covering most of the microfluidic chip due to having twice as much representation in the pattern with $20$ seconds of injection compared to red or green dyes each having only ten seconds.

\begin{figure}[H]
\centering
\includegraphics[width=1\textwidth]{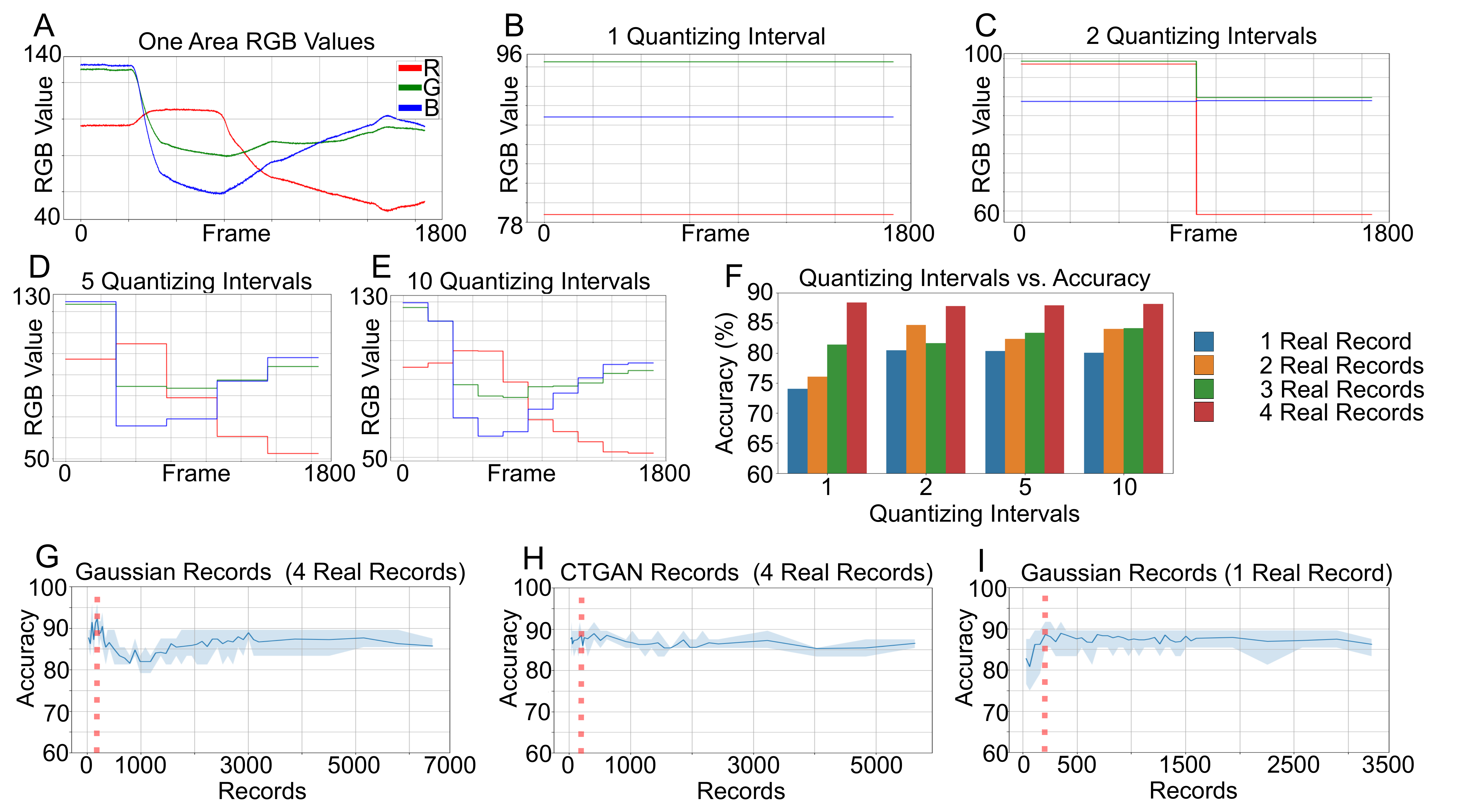}
\caption{(A) One example output of the microfluidic chip showing RGB values.
(B,C,D,E) Quantized versions of the previous graph, showing how each range is averaged over $1$, $2$, $5$, and $10$ intervals.
(F) An evaluation of how quantizing intervals and real records affects accuracy, with real records being per pattern.
(G) An analysis on how Gaussian data generation with $32$ real records, four per pattern, performs at different record totals using five quantized intervals. There is a red dotted line at 200 records.
(H) Finding optimal record count using Conditional Tabular Generative Adversarial Network (CTGAN) data generation with $32$ real records, four per pattern, used to generate data using five quantized intervals. There is a red dotted line at 200 records.
(I) Similar to the previous Gaussian data generation graph, this graph only uses eight real records, one per pattern, to generate synthetic records and test optimal record counts. There is a red dotted line at 200 records.
}
\label{fig:Figure4}
\end{figure}

\subsection{Pre-Processing Data and Readout Layer}

\label{sec:Pre-Processing Data and Readout Layer}

The microfluidic chip in our system serves as the core computational reservoir, with the output being a temporal signal. This raw reservoir output needs to be transformed from the temporal domain into the spatial domain to be used as features in the readout layer. We use quantizing as a strategy to extract, pre-process, and condense high-dimensional temporal data into a compact set of features for classification. Quantization takes each of the nine raw output signals from the reservoir and averages the values over specific intervals to produce features, as further explained in Section~\ref{sec:Data Quantization and Train-Test Split}. After quantizing, the reservoir output signals, the data is split into a training and testing spatio-dataset with $32$ real records in the training set and $48$ real records in the test set. Then, the spatio-dataset is standardized using a global scalar normalization to center and scale the data uniformly. Next, we design a readout layer implemented in Python on the silicon side of our system and train a model to verify that this method of representing data performs well for our system. To ensure consistency, each model is trained $50$ times with randomly initialized weights using the same training data, with the average, minimum, and maximum accuracies of these runs preserved for later analysis. After verifying model performance, we then generate quantized synthetic outputs (synthetic records) to further improve the performance of the model, as described in Section~\ref{sec:Increasing Accuracy with Synthetic Data}. Synthetic data can benefit machine learning when applied to datasets to handle privacy concerns\cite{rajotte_synthetic_2022}, increase dataset variability\cite{jacobsen_machine_2023}, and increase model performance with limited datasets\cite{lu_machine_2023}. Our use case for synthetic data is to expand our limited dataset and save time on data collection\cite{de_melo_next-generation_2022}.

\subsubsection{Data Quantization and Train-Test Splitting}
\label{sec:Data Quantization and Train-Test Split}

Data preparation and processing are critical to the performance of our system, which we thoroughly analyze in this section. When quantizing data from the microfluidic chip, each of the nine time series is divided into uniform intervals, and the average output value is calculated within each interval. This approach preserves overall reservoir output trends, enables spatio-temporal conversion, and significantly reduces the number of features passed to the model. For example, dividing each reservoir output into ten intervals reduces the dimensionality from $16,200$ to only $90$ features(for details on readout layer size, refer to Supplementary Material Note S1).  Figure~\ref{fig:Figure4}(A) shows the raw reservoir output of a single experiment, while Figure~\ref{fig:Figure4}(B–E) depict quantized versions using $1$, $2$, $5$, and $10$ intervals, respectively. Although finer quantization provides greater temporal resolution, even coarse quantization retained distinctive reservoir output structures that supported accurate classification, which will be discussed later in our performance metrics. After applying quantization, the spatio-dataset is split into a training set containing $32$ real records and a testing set containing $48$ real records. The $48$ testing records are never modified, while the $32$ training records are varied by including $8$, $16$, $24$ or $32$ in the spatio-dataset for training.

To evaluate the effect of quantization granularity, we train multiple models using intervals of $1$, $2$, $5$, and $10$ for quantizing. As shown in Figure~\ref{fig:Figure4}(F), classification accuracy remains high across all quantization intervals, reaching around $88\%$ at each interval. On the same graph, we also test how varying the number of records we use for training affects accuracy. RC architectures are inherently well-suited for low-data scenarios due to their fixed internal dynamics and lightweight readout layers, making them ideal for our limited-data setup. The four colored bars in the graph represent using one, two, three or four training records per pattern (or $8$, $16$, $24$, and $32$ total training records). With fewer records, the accuracy drops across all quantization intervals, with the lowest being $74\%$ at one quantization interval. In particular, a $1$-quantization interval produces a broader accuracy range, with lows below $75\%$, while a $10$-quantization interval shows a narrower range and a minimum accuracy of $80\%$. These differences are attributed to the number of training records and the sizes of the quantization interval. Low-resolution quantization results in a smaller readout layer, whereas higher-resolution quantization, even with limited data, captures more temporal detail, with the trade-off being a larger readout layer. These results underscore the effectiveness of quantizing data as a scalable and practical method for reducing dimensionality while maintaining strong model performance. In the same graph, performance plateaus after 4 records per pattern, suggesting that relatively few physical experiments are sufficient to capture the essential input–output behavior of the system (further details in training models with varying quantization intervals and records can be found in Supplementary Materials Note S2). In the next section, we analyze the accuracy performance when we integrate synthetic records to the training dataset.

\subsubsection{Increasing Accuracy with Synthetic Data}
\label{sec:Increasing Accuracy with Synthetic Data}

In order to overcome the lack of training data, we expand the training dataset with the addition of synthetic records. In this study, we focus on two data generation methods, applying Gaussian\cite{becerra-suarez_improvement_2025, goldt_gaussian_2020} noise to quantized reservoir outputs and training a Conditional Tabular Generative Adversarial Network (CTGAN)\cite{xu_modeling_2019} model to generate more records based on quantized reservoir outputs (details on the Gaussian and CTGAN data generation are located in Supplementary Material Note S3). These modifications simulate variability and expand the dataset while retaining the essential characteristics of the original measurements. An important note is that synthetic records are only ever used for training, and never used in the test set. After quantization and before incorporating synthetic data, the real records are split, with $32$ real records set aside for training and $48$ real records set aside for testing.

When studying synthetic data generation, we investigate the performance of each data generation method against the number of total data records utilized. Figure~\ref{fig:Figure4}(G-I) compare CTGAN and Gaussian-based datasets across increasing record totals with a maximum of $6500$ to find which synthetic data generation record count is the most efficient. Figure~\ref{fig:Figure4}(G) uses Gaussian noise applied to four quantized reservoir outputs per pattern to generate data, resulting in the highest performance at $200$ total records. At $200$ total records, there is a short spike in accuracy, a $92\%$ average, when a small number of records is used, which then drops almost immediately. A similar behavior is observed in Figure~\ref{fig:Figure4}(H), where CTGAN is used to generate records. In this case, the improvement is small and remains below $90\%$ and the accuracy range is also minimal, again around $200$ records. Although both approaches initially benefit the model, performance gains diminish as the total number of training records increases. This diminishing behavior is evident in settings that use higher quantization intervals and full coverage of the output regions, where the original feature set already captures substantial reservoir output variability.

When the input is constrained for further testing, as in the case of Figure~\ref{fig:Figure4}(I), with two quantization intervals and only one quantized reservoir output per pattern used to generate the synthetic dataset, data augmentation plays a more critical role. Under these conditions, increasing the total number of synthetic records provides measurable improvements, especially when working with limited information per record. There is an increase between $10\%$ and $15\%$ between the minimum and maximum accuracy when the real records are combined with synthetic records, making a total of $200$ records. Although the microfluidic reservoir system performs well with a modest number of real records, synthetic data generation, particularly in the form of Gaussian noise augmentation, proves essential in scaling the performance under constrained or resource-limited testing conditions. This need for synthetic data could also arise when the complexity of a dataset increases or when the application of the trained model is of critical importance.

\subsection{Dimensionality Reduction of Readout Layer}
\label{sec:Dimensionality Reduction of Readout Layer}

To improve the efficiency of our RC system, we employ dimensionality reduction techniques by pruning the reservoir outputs, thereby reducing the size of the readout layer. Although using a larger number of input features can increase classification accuracy, it also increases computational cost and model complexity. To optimize the readout layer, we first evaluate the MI between the input patterns and each output area to identify the most informative regions. We then systematically test different combinations of output areas and quantization intervals to determine configurations that offer the best trade-off between classification accuracy and architectural simplicity.

\begin{figure}[H]
\centering
\includegraphics[width=1\textwidth]{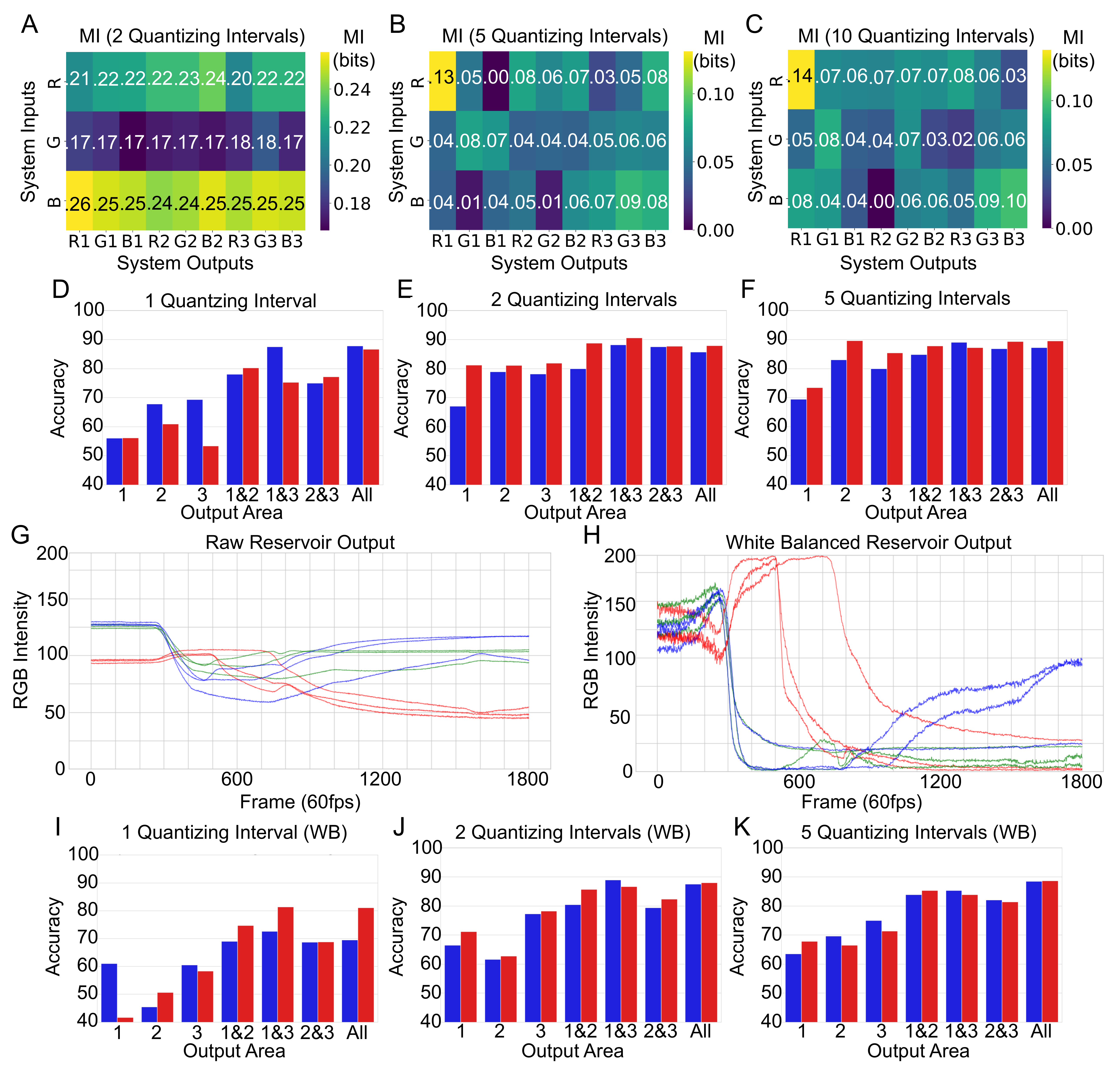}
\caption{
(A–C) MI between the microfluidic chip's three inputs and nine outputs, with the outputs represented as R, G, and B over three areas. This was all analyzed over two, five, and ten quantization intervals.
(D–F) Accuracy comparison using different chip output areas, with quantization intervals of $1$, $2$, and $5$.
(G) Raw reservoir output as captured by the camera.
(H) White-balanced (WB) data illustrating the effect on reservoir output signal representation.
(I-K) Same as D–F, but using WS data for model training.
}
\label{fig:Figure5}
\end{figure}

\subsubsection{Mutual Information of Reservoir Inputs and Outputs}
\label{sec:Mutual Information of Reservoir Inputs and Outputs}

We start the process of pruning the reservoir outputs by analyzing MI between the input patterns, represented as $3 \times 5$ binary matrices with $1$ if the pulse is on and $0$ when there is no pulse in the injection time slot, and the RGB reservoir output values recorded in each of the three output areas of the chip after quantization. This analysis allows us to quantify how informative each output area is with respect to the original input, under different quantization settings. Figure~\ref{fig:Figure5}(A–C) shows the MI for the quantization intervals $2$, $5$ and $10$. Further information on the MI calculation is located in supplementary note S4.

For the two quantization intervals shown in Figure~\ref{fig:Figure5}(A), the red and blue channels consistently show strong contributions in all areas with MI values around $0.22$ bits and $0.25$ bits respectively, while the green channel contains less information at $0.17$ bits. This aligns with the physical layout of the chip in Figure~\ref{fig:Figure3}(A), where area $2$ that is dominated by green dye lies between the red and blue injection paths and is most susceptible to mixing effects. While this data shows overall trends, detail is lost with each cell being around the same value. Figure~\ref{fig:Figure5}(B) shows quantization at five intervals. In this experiment, the red input with the red output in area $1$ has the highest MI at $0.13$ bits and there is a spread of MI throughout the heatmap with only three areas having an MI value of $0$ bits. At the highest resolution of ten quantization intervals in Figure~\ref{fig:Figure5}(C), a measurable amount of MI is in most cells with the red input and output in area $2$ having the highest MI at $0.14$ bits and the blue input with the red output of area $2$ being the only cell with an Mi value of $0$ bits.

There is a measurable amount of MI between all output columns and the patterns input to the system, making it difficult to exclude entire output areas. As quantization resolution decreases in the system, behavior such as blue having a higher impact in our system than red or green can be observed. Quantization at $2$ intervals does not have the finer details captured by higher quantization intervals with similar values carried between all inputs and outputs. When quantization resolution increases, the details of specific signals increase, leading to areas such as the red output in area $1$ showing greater impact with MI values of $0.13$ bits and $0.14$ bits. This helps us to understand the pattern of information flowing through the system, where different areas can give a MI value between $0$ to $0.14$ bits. Pruning output areas from the training must be approached carefully, as informative signals can still be present in weaker-performing areas depending on the color channel. For an analysis on specific pattern MI, please refer to Supplementary Material Notes S5 - S7.

\subsubsection{Impact of Output Area Selection on System Performance}
\label{sec:Impact of Variations in Output Areas}

To translate the MI observations into practical improvements, we test multiple combinations of reservoir output areas under varying quantization resolutions. Each configuration is evaluated across $50$ independent training iterations to account for stochastic variation. Figures~\ref{fig:Figure5}(D–F) show the results of selecting different output areas to train and test models. The red bars represent models trained with Gaussian synthetic records: $32$ real records combined with $168$ synthetic records, totaling $200$. The blue bars use CTGAN to generate the same number of synthetic records for training. To evaluate whether certain microfluidic reservoir outputs carry more or less information for classification, the plots show the accuracy when training with different output areas, specifically, area $1$, $2$, $3$, or various combinations of these, for both CTGAN and Gaussian synthetic data. In Figure~\ref{fig:Figure5}(D), the system is quantized in a single interval, where the entire output space is treated uniformly. With only one output region used for training, the accuracy is low, ranging from $55\%$ to $70\%$. Individual output areas $1$, $2$, and $3$, each contain only three input nodes ($24$ edges) in the readout layer, which is insufficient to represent the full dynamics of the system. When using two output areas, the accuracy improves. In particular, combining output areas $1$ and $3$ with CTGAN-synthetic records yields an accuracy of $87\%$ using only $48$ edges in the read-out layer. Interestingly, including all $3$ output areas does not further increase accuracy, which plateaus around $88\% $. This trend continues in the other quantization settings. In Figure~\ref{fig:Figure5}(E), two quantization intervals are used, doubling the number of edges trained and leading to higher overall accuracy. When training in a single output region, the Gaussian data outperforms the CTGAN data, with output area $3$ achieving $82\% $ accuracy. This trend persists when using two output areas: areas $1$ and $3$ reach $91\% $ accuracy using Gaussian records. Once again, incorporating all three output areas does not improve performance. Finally, Figure~\ref{fig:Figure5}(F) shows results using five quantization intervals. Training on a single output area results in a readout layer with $120$ edges. Using only output area $2$ achieves $89\% $ accuracy. However, when training using output area $1$, performance decreases with Gaussian synthetic data. Adding more output areas (two or three total) does not noticeably improve accuracy compared to previous tests.

In all trials, reducing the number of output areas leads to fewer input features and lower model complexity. Although this sometimes reduces accuracy, the output area combination $1$ and $3$ performs consistently well, especially with two quantization intervals. In both the $1$ and $2$ interval cases, training with area $1$ and $3$ outperforms using output area $2$ alone. This trend is further supported in the next set of experiments when a white-balancing process is applied to the experimental videos. In summary, the highest average accuracy observed is $91\%$, achieved using $2$ quantization intervals, output areas $1$ and $3$, $32$ real records, and $168$ Gaussian-synthesized records.

\subsubsection{Incorporating White Balancing}
\label{sec:Incorporating White Balancing}

During video-based data collection, the raw outputs extracted from the microfluidic reservoir range between $40$ and $120$, as shown in Figure~\ref{fig:Figure5}(G). This restricted range limits the visual contrast of the data and makes it challenging to distinguish signal dynamics, especially in lower-resolution patterns.

To address this challenge, we apply a white balancing technique using the OpenCV Python library’s \emph{xphoto.createSimpleWB} function, which operates under the \emph{gray-world} assumption by adjusting the colors in each frame so that the overall image appears gray. As shown in Figure~\ref{fig:Figure5}(H), this process effectively extends the visible RGB signal range between $0$ and $200$, improving contrast and reducing the impact of lighting variability. Although white balancing significantly improves the visual quality of reservoir outputs for the human eye, it does not improve the performance of trained models. In fact, models trained on standardized data under perform compared to those trained on unprocessed reservoir outputs, as shown in Figure~\ref{fig:Figure5}(I-K). In Figure~\ref{fig:Figure5}(I), accuracy across the graph never exceeds $82\%$, while in Figure~\ref{fig:Figure5}(D), accuracies reach $87\%$. This trend can also be observed in Figure~\ref{fig:Figure5}(J-K) where the accuracy never exceeds $90\%$. Therefore, white balancing of the videos is only beneficial for data in a visualization application. Based on this the optimal classification results reported in this work are obtained using raw (non-standardized) reservoir outputs. Following previously described trends, including output area $2$ in training usually results in a decrease in accuracy compared to using output area $1$ and $3$. Despite the limitations in lighting and signal contrast, the unprocessed data is robust enough to support classification accuracies of up to $91\%$, demonstrating the resilience of the microfluidic reservoir.

\subsubsection{Analyzing Classification Errors}
\label{sec:Pattern Analysis}

\begin{figure}[H]
\centering
\includegraphics[width=1\textwidth]{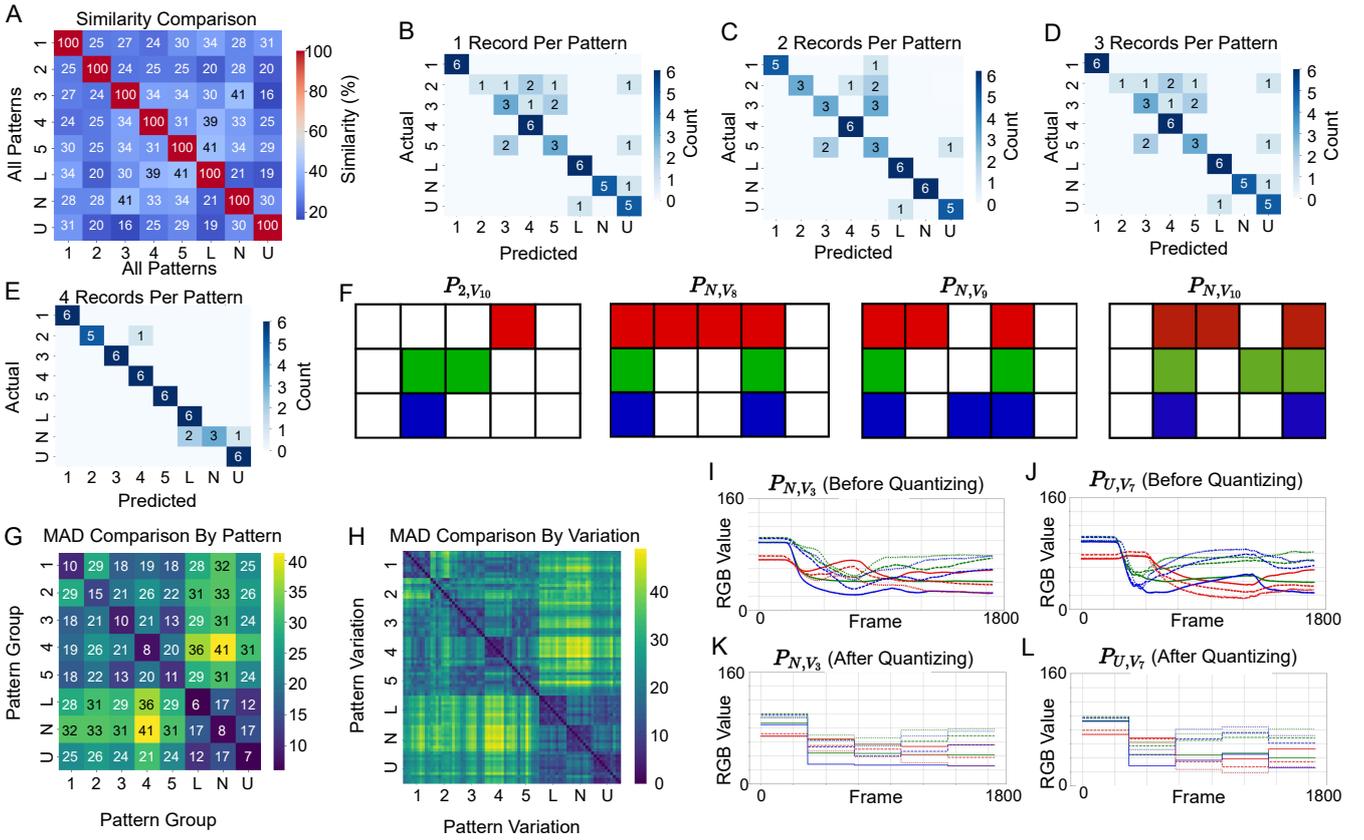}
\caption{
(A) Similarity between input patterns before entering the system.
(B) Confusion matrix from a model trained using $200$ total records, $8$ real records and $192$ synthetic records.
(C) Confusion matrix from a model trained using $200$ total records, $16$ real records and $184$ synthetic records.
(D) Confusion matrix from a model trained using $200$ total records, $24$ real records and $176$ synthetic records.
(E) Confusion matrix from the highest-performing model trained using $200$ total records, $32$ real records and $168$ synthetic records.
(F) Four patterns that got classified incorrectly in the best performing model.
(G) Heatmap showing the mean absolute difference (MAD) between patterns after passing through the reservoir. This is measured through the difference in RGB values and is represented as an RGB value.
(H) Heatmap showing variation within each pattern after passing through the reservoir being compared through a MAD value.
(I,J) Two raw reservoir output examples of patterns that appeared highly similar in the difference heatmap.
(K,L) Two quantized reservour outputs of patterns that appeared highly similar in the difference heatmap.
}
\label{fig:Figure6}
\end{figure}

Although the system achieves consistently high performance, with classification accuracies around $91\%$, it does not reach the $100\%$ benchmark that could be expected for a relatively restricted pattern recognition task. To investigate the source of remaining errors, we conducted an in-depth analysis of input data focused on signal separability and pattern misclassifications. Figure~\ref{fig:Figure6}(A) shows a similarity matrix of the original input patterns prior to entering the reservoir. The most pronounced similarity occurs between patterns $\bf{P_{3}}$, $\bf{P_{4}}$, $\bf{P_{5}}$, $\bf{P_{N}}$ and $\bf{P_{L}}$, where patterns $\bf{P_{3}}$ and $\bf{P_{N}}$ have a $41\%$ match, patterns $\bf{P_{4}}$ and $\bf{P_{L}}$ have a $39\%$ match, and patterns $\bf{P_{5}}$ and $\bf{P_{L}}$ have a $41\%$ match. These similarities are due to the ten variations of each pattern, with certain variations being more similar than others. After comparing the patterns used as inputs to the system, we examine which patterns are misclassified in different models. The first trial uses one real record per pattern and $192$ Gaussian synthetic records with the resulting confusion matrix in Figure~\ref{fig:Figure6}(B). In this confusion matrix, patterns $\bf{P_{2}}$ and $\bf{P_{3}}$ are frequently misclassified. As real records increase, reaching four real records per pattern and $168$ Gaussian synthetic records, the performance of the model improves significantly (Figure~\ref{fig:Figure6}E). At this stage, all patterns achieve perfect classification except $\bf{P_{2}}$ and $\bf{P_{N}}$.

To understand this residual error, we isolate misclassified pattern variations and analyze them separately (Figure~\ref{fig:Figure6}(F)). Three of these variations are part of pattern $\bf{P_{N}}$ while one is pattern $\bf{P_{2}}$. This high representation of $\bf{P_{N}}$ in misclassifications is explained in the next test. The heat maps of raw reservoir output differences (Figure~\ref{fig:Figure6}(G)) reveal that patterns $\bf{P_{U}}$, $\bf{P_{N}}$, and $\bf{P_{L}}$ produce highly similar signatures, with pattern $\bf{P_{2}}$ showing the highest internal variability. The difference comparison is obtained by calculating the mean absolute difference between patterns, with findings that match the patterns that kept getting misclassified,. The MAD comparison by variation (Figure~\ref{fig:Figure6}(H)) further emphasize the difficulty in distinguishing between patterns with minimal signal divergence. In this heat map, patterns $\bf{P_{U}}$, $\bf{P_{N}}$ and $\bf{P_{L}}$ have a higher similarity compared to patterns $\bf{P_{4}}$ and $\bf{P_{N}}$, which have a lower similarity (a deeper analysis on similarities in Figure~\ref{fig:Figure6}(H) is in Supplementary Material Note S8). As an example that can lead to misclassification, a direct comparison of signals from patterns $\bf{P_{N}}$ and $\bf{P_{U}}$ (Figure~\ref{fig:Figure6}(I,J)) shows that the two patterns follow similar trends with red decreasing, while blue and green increase. After quantization, as shown in Figure~\ref{fig:Figure6} (K,L), the patterns appear even more similar. Therefore, if one of these patterns is used to train the readout layer, then the other pattern is likely to get misclassified. For a closer evaluation of mean absolute differences between patterns and their variations, refer to Supplementary Material Note S9.

These findings point to a fundamental limitation of the current system: its low spatio-temporal resolution. When two input patterns produce nearly identical quantized reservoir outputs, either due to overlap in structure, mixing effects, or quantization compression, the model lacks the signal diversity needed to reliably separate them. Enhancing system resolution through either more refined injection control or finer-grained output sensing will likely improve the performance and expand the number of patterns that can be classified with confidence.

\section{Discussion and Conclusion}
\label{sec:Discussion and Conclusion}

This study presents a novel microfluidic RC device that uses the dynamics of fluid flow through a biologically inspired structure for pattern classification tasks. The varying points of injection, interactions between channel intersections, and the timing of pattern injection all lead to dye mixing suitable for reservoir computing. Drawing from the natural vein architecture of a dragonfly wing, the system operates as a physical reservoir, passively encoding temporal patterns into color-coded fluid signatures. In the future, other systems such as liquid channels in plants, organs in organisms, and naturally occurring dynamic environments could all be modeled as nature-inspired computing mediums. The signals processed through the reservoirs inspired by these ideas can then be passed to a lightweight software-based readout layer that classifies the signals for various tasks.

Our experiments demonstrate that even with coarse resolution and simple RGB-based detection, the system reliably transforms structured reservoir inputs into distinguishable reservoir outputs. With only $32$ real records and minimal pre-processing, our model consistently achieves classification accuracies of up to $88\%$. This performance is further improved through the addition of synthetically generated records, particularly using Gaussian noise, to achieve an accuracy of $91\%$ using all $80$ pattern variations. This approach proves effective, efficient, and well suited to the small sample quantity of our experiments. Generating a Gaussian noise-modified dataset is efficient, as it does not require model training for each pattern to produce new records. This approach also reduces sample collection time, mitigates overfitting\cite{du_synaptic_2022}, and mimics the reservoir output signal changes inherent in our RGB detection method by adjusting the raw reservoir output data. A key strength of this system lies in its simplicity. Unlike traditional neural networks, which require extensive training and computational resources, our microfluidic reservoir does not require learning within its core structure. The dynamics of the fluid network itself provide the non-linearity and memory, while training is confined to the final readout layer. This not only reduces the training burden, but also enables rapid prototyping and evaluation.

However, the system is not without limitations. The primary constraint is resolution, both in spatial pattern encoding and in temporal signal capture. Using modulation techniques, changing the flow rate, or encoding information in a unique way can possibly increase the resolution of the reservoir inputs in our system. Our use of RGB dye and a fixed $3\times 5$ input grid limits the expressiveness of the input space. Furthermore, quantization, while necessary to reduce dimensionality, sometimes compresses distinguishing features between similar reservoir outputs. This was evident in pattern analysis, where the visual similarity between patterns $\bf{P_{N}}$ and $\bf{P_{U}}$ led to persistent misclassifications. Additionally, the reliance on passive retention of dye means that signals persist unless flushed, which may limit re-usability or real-time adaptability in future deployments. Future work should explore options for reservoir channel clearing and additional system feedback to address this static behavior. Lighting conditions and hardware constraints such as manually turning the pumps on and off and a clear system that detects high RGB signals further introduce noise and require consideration when performing tests. While white balancing improves visualization, it does not improve classification and is ultimately excluded from model training, with accuracies in every quantization interval below or on par with raw reservoir output tests, never exceeding $90\%$ accuracy. 

Despite these challenges, the microfluidic architecture shows strong potential for an alternative form of RC. Future work should explore increased spatial resolution, additional inlet ports, and alternative sensing modalities (e.g., chemical, capacitive or fluorescence-based detection). More advanced microfluidic fabrication could also enable branching or feedback loops within the reservoir, further expanding its dynamic range and computational capability.

\section{METHODS}
\label{sec:Methods}

\subsection{Readout Layer of the System}
\label{sec:Readout Layer of the System}

The primary computational complexity of the system lies within the microfluidic chip, leaving the readout layer to be implemented in Python. The readout layer consists of a single dense layer, taking between $3$ and $45$ input features and producing an output across eight patterns. To verify the consistency of the model, $50$ independent models were trained on the same dataset and their average, minimum, and maximum accuracies were recorded. Before training, the spatio-dataset was standardized using a single scalar value applied uniformly throughout the data set. The output labels were one-hot encoded for compatibility with the categorical classification model. After testing various configurations, it was found that the optimal training setup used the Adam optimizer with a learning rate of $0.02$, and a softmax activation function in the final layer. Training was conducted for up to $300$ epochs, with early stopping implemented to prevent overfitting.

\subsection{Experimental Setup}
The experimental workflow is divided into three main stages: injection, detection, and data processing. Each stage is carefully designed to ensure the accurate generation, capture, and interpretation of microfluidic patterns within the microfluidic RC platform.

\subsubsection{Injection Phase}
Input patterns are constructed as $3\times5$ binary grids, with each row corresponding to a primary color: red (top), green (middle), and blue (bottom). A binary $1$ within a cell indicates that dye is injected into the corresponding color channel for a duration of five seconds (equal to 300 video frames), while a $0$ denotes no flow injection. These patterns simulate simple spatio-temporal structures to be encoded by the microfluidic reservoir. Three Chemyx F100T2 syringe pumps, each loaded with a $10 mL$ syringe, inject red, green, and blue food coloring (McCORMICK Culinary and Chef-O-Van) into the chip’s inlet ports. The flow rate is set to $3 mL/min$ per inlet. Flexible Tygon plastic tubing (inner diameter: $0.02 in$; outer diameter: $0.06 in$) connects the syringes to the microfluidic device.

Each test begins with the microfluidic chip filled with clear water. The pumps attached to the chip are manually activated according to the encoded pattern and after each test is complete, the chip is flushed with clear water to remove any residual dye. To perform this flush, the three pumps are reloaded with water and operated simultaneously. This step ensures that no color is carried over between experiments and that the reservoir is reset to a neutral state. In cases where air bubbles become trapped within the microfluidic channels, potentially blocking flow and altering dynamics, a hydrostatic clearing method is employed (this is presented in Supplementary Note S10). Water-filled tubing is connected to both the inlet and outlet ports, with the water source elevated above the chip. The resulting hydrostatic pressure forces air through the \emph{polydimethylsiloxane} (PDMS) material.

\subsubsection{Detection Phase}
During each test, the entire microfluidic chip is filmed using a Phantom Miro M310 high-speed camera. The camera is positioned directly above the device and records at $60 fps$ for a total duration of $30$ seconds. This time frame captures the full propagation of the dye patterns from the inlets to the outlets. Following image acquisition, video frames are processed using a custom Python script using the OpenCV library. Three detection areas are manually selected in the diamond-shaped areas at the bottom of the chip in Figure~\ref{fig:Figure3}(A), where color mixing and retention occurs. RGB values are extracted from each selected region across all frames, producing nine total signals (three areas $\times$ three color channels) per experiment.

\subsubsection{Processing Phase}
To reduce the dimensionality of the input and transform the microfluidic reservoir outputs from temporal data to spatio-temporal features in preparation for classification, each of the nine time series ($1800$ points per signal) is quantized. Depending on the experiment, these signals are divided between $1$ and $10$ temporal intervals, and the average RGB value within each interval is computed. This quantization reduces the feature count from $16,200$ to a manageable range of $3$ to $90$ features, depending on the number of areas and quantization intervals we use. The resulting feature vectors are split into a training dataset of $32$ records and testing dataset of $48$ records, which are then passed to a Python-based dense readout layer, which performs classification across eight possible input patterns. Model training and inference were performed using standard machine learning workflows, with accuracy tracked across repeated trials.

\subsubsection{Microfluidic Chip Design}

As described in Ryu et al. \cite{ryu_insect_2025}, section two, the microfluidic chip was made using a $1:1$ vein mold of the forewing of the \emph{Common Green Darner} dragonfly. The device was created through soft lithography on a 3D printed master mold. The master mold was designed in SolidWorks by taking a high resolution image of the wing, converting it to binary, and importing it into CAD. There, the wing was reconstructed and flow ports were added to areas around the wing. The mold was then printed using a stereo-lithography 3D printer (MiiCraft Ultra 100, Creative CADWorks3D) and processed in the steps outlined by Emeigh et al.\cite{emeigh_side-view_2025}. The mold was then used to make the wing device by pouring PDMS (Sylgard 184, Dow-Corning) over the mold and curing in a $60^{\circ}C$ oven overnight. Using biopsy punches (Rapid Core Punch, World Precision Instruments), holes were punched in the PDMS body to act as flow connection ports into the system. Subsequently, the device was bonded to a glass slide using a plasma cleaner (PDC-001, Harrick Plasma) and placed into an $80^{\circ}C$ oven overnight for bond permanence. After the device was removed from the oven, tubing was inserted into the flow ports completing the construction of the device.

\bibliography{bibliography}

\begin{thebibliography}{58}
\providecommand{\natexlab}[1]{#1}
\providecommand{\url}[1]{\texttt{#1}}
\expandafter\ifx\csname urlstyle\endcsname\relax
  \providecommand{\doi}[1]{doi: #1}\else
  \providecommand{\doi}{doi: \begingroup \urlstyle{rm}\Url}\fi

\bibitem[Angerbauer et~al.(2024{\natexlab{a}})Angerbauer, Enzenhofer, Pankratz, Hamidovic, Springer, and Haselmayr]{angerbauer_novel_2024}
Stefan Angerbauer, Franz Enzenhofer, Tobias Pankratz, Medina Hamidovic, Andreas Springer, and Werner Haselmayr.
\newblock Novel {Nano}-{Scale} {Computing} {Unit} for the {IoBNT}: {Concept} and {Practical} {Considerations}.
\newblock \emph{IEEE Transactions on Molecular, Biological, and Multi-Scale Communications}, 10\penalty0 (4):\penalty0 549--565, December 2024{\natexlab{a}}.
\newblock ISSN 2372-2061, 2332-7804.
\newblock \doi{10.1109/TMBMC.2024.3397050}.
\newblock URL \url{https://ieeexplore.ieee.org/document/10534193/}.

\bibitem[Angerbauer et~al.(2024{\natexlab{b}})Angerbauer, Pankratz, Enzenhofer, Springer, Khanzadeh, and Haselmayr]{angerbauer_molecular_2024}
Stefan Angerbauer, Tobias Pankratz, Franz Enzenhofer, Andreas Springer, Roya Khanzadeh, and Werner Haselmayr.
\newblock Molecular {Nano} {Neural} {Networks} ({M3N}): {In}-{Body} {Intelligence} for the {IoBNT}.
\newblock In \emph{{ICC} 2024 - {IEEE} {International} {Conference} on {Communications}}, pages 4819--4824, Denver, CO, USA, June 2024{\natexlab{b}}. IEEE.
\newblock ISBN 978-1-7281-9054-9.
\newblock \doi{10.1109/ICC51166.2024.10622358}.
\newblock URL \url{https://ieeexplore.ieee.org/document/10622358/}.

\bibitem[Ansarizadeh et~al.(2025)Ansarizadeh, Nguyen, Lazovic, Kettunen, De~Silva, Sivakumar, Junttila, Rissanen, Hicks, Singh, and Eklund]{ansarizadeh_microfluidic_2025}
Mohammadhassan Ansarizadeh, Hoang-Tuan Nguyen, Bojana Lazovic, Jere Kettunen, Laknee De~Silva, Ragul Sivakumar, Pauliina Junttila, Siiri-Liisa Rissanen, Ryan Hicks, Prateek Singh, and Lauri Eklund.
\newblock Microfluidic vessel-on-chip platform for investigation of cellular defects in venous malformations and responses to various shear stress and flow conditions.
\newblock \emph{Lab on a Chip}, 25\penalty0 (4):\penalty0 613--630, 2025.
\newblock ISSN 1473-0197, 1473-0189.
\newblock \doi{10.1039/D4LC00824C}.
\newblock URL \url{https://xlink.rsc.org/?DOI=D4LC00824C}.

\bibitem[Azizbeigi et~al.(2021)Azizbeigi, Zamani~Pedram, and Sanati-Nezhad]{azizbeigi_microfluidic-based_2021}
Kasra Azizbeigi, Maysam Zamani~Pedram, and Amir Sanati-Nezhad.
\newblock Microfluidic-based processors and circuits design.
\newblock \emph{Scientific Reports}, 11\penalty0 (1):\penalty0 10985, May 2021.
\newblock ISSN 2045-2322.
\newblock \doi{10.1038/s41598-021-90485-z}.
\newblock URL \url{https://www.nature.com/articles/s41598-021-90485-z}.

\bibitem[Bai et~al.(2023)Bai, Thiem, Lombardi, Liang, and Yi]{bai_design_2023}
Kang~Jun Bai, Clare Thiem, Jack Lombardi, Yibin Liang, and Yang Yi.
\newblock Design {Strategies} and {Applications} of {Reservoir} {Computing}: {Recent} {Trends} and {Prospects} [{Feature}].
\newblock \emph{IEEE Circuits and Systems Magazine}, 23\penalty0 (4):\penalty0 10--33, 2023.
\newblock ISSN 1531-636X, 1558-0830.
\newblock \doi{10.1109/MCAS.2023.3325496}.
\newblock URL \url{https://ieeexplore.ieee.org/document/10379007/}.

\bibitem[Becerra-Suarez et~al.(2025)Becerra-Suarez, Alvarez-Vasquez, and Forero]{becerra-suarez_improvement_2025}
Fray~L. Becerra-Suarez, Halyn Alvarez-Vasquez, and Manuel~G. Forero.
\newblock Improvement of {Bank} {Fraud} {Detection} {Through} {Synthetic} {Data} {Generation} with {Gaussian} {Noise}.
\newblock \emph{Technologies}, 13\penalty0 (4):\penalty0 141, April 2025.
\newblock ISSN 2227-7080.
\newblock \doi{10.3390/technologies13040141}.
\newblock URL \url{https://www.mdpi.com/2227-7080/13/4/141}.

\bibitem[Bhatia and Ingber(2014)]{bhatia_microfluidic_2014}
Sangeeta~N Bhatia and Donald~E Ingber.
\newblock Microfluidic organs-on-chips.
\newblock \emph{Nature Biotechnology}, 32\penalty0 (8):\penalty0 760--772, August 2014.
\newblock ISSN 1087-0156, 1546-1696.
\newblock \doi{10.1038/nbt.2989}.
\newblock URL \url{https://www.nature.com/articles/nbt.2989}.

\bibitem[Bradski(2000)]{opencv_library}
G.~Bradski.
\newblock {The OpenCV Library}.
\newblock \emph{Dr. Dobb's Journal of Software Tools}, 2000.

\bibitem[Cai et~al.(2023)Cai, Ao, Tian, Wu, Liu, Tchieu, Gu, Mackie, and Guo]{cai_brain_2023}
Hongwei Cai, Zheng Ao, Chunhui Tian, Zhuhao Wu, Hongcheng Liu, Jason Tchieu, Mingxia Gu, Ken Mackie, and Feng Guo.
\newblock Brain organoid reservoir computing for artificial intelligence.
\newblock \emph{Nature Electronics}, 6\penalty0 (12):\penalty0 1032--1039, December 2023.
\newblock ISSN 2520-1131.
\newblock \doi{10.1038/s41928-023-01069-w}.
\newblock URL \url{https://www.nature.com/articles/s41928-023-01069-w}.

\bibitem[Canaday et~al.(2018)Canaday, Griffith, and Gauthier]{canaday_rapid_2018}
Daniel Canaday, Aaron Griffith, and Daniel~J. Gauthier.
\newblock Rapid time series prediction with a hardware-based reservoir computer.
\newblock \emph{Chaos: An Interdisciplinary Journal of Nonlinear Science}, 28\penalty0 (12):\penalty0 123119, December 2018.
\newblock ISSN 1054-1500, 1089-7682.
\newblock \doi{10.1063/1.5048199}.
\newblock URL \url{https://pubs.aip.org/cha/article/28/12/123119/135920/Rapid-time-series-prediction-with-a-hardware-based}.

\bibitem[Chembo(2020)]{chembo_machine_2020}
Yanne~K. Chembo.
\newblock Machine learning based on reservoir computing with time-delayed optoelectronic and photonic systems.
\newblock \emph{Chaos: An Interdisciplinary Journal of Nonlinear Science}, 30\penalty0 (1):\penalty0 013111, January 2020.
\newblock ISSN 1054-1500, 1089-7682.
\newblock \doi{10.1063/1.5120788}.
\newblock URL \url{https://pubs.aip.org/cha/article/30/1/013111/1027406/Machine-learning-based-on-reservoir-computing-with}.

\bibitem[Chen et~al.(2024)Chen, Wang, Yang, Chen, Zhang, and Zeng]{chen_full-analog_2024}
Liangyu Chen, Xiaoping Wang, Chao Yang, Zhanfei Chen, Junming Zhang, and Zhigang Zeng.
\newblock Full-{Analog} {Reservoir} {Computing} {Circuit} {Based} on {Memristor} {With} a {Hybrid} {Wide}-{Deep} {Architecture}.
\newblock \emph{IEEE Transactions on Circuits and Systems I: Regular Papers}, 71\penalty0 (2):\penalty0 501--514, February 2024.
\newblock ISSN 1549-8328, 1558-0806.
\newblock \doi{10.1109/TCSI.2023.3334267}.
\newblock URL \url{https://ieeexplore.ieee.org/document/10330156/}.

\bibitem[De~Melo et~al.(2022)De~Melo, Torralba, Guibas, DiCarlo, Chellappa, and Hodgins]{de_melo_next-generation_2022}
Celso~M. De~Melo, Antonio Torralba, Leonidas Guibas, James DiCarlo, Rama Chellappa, and Jessica Hodgins.
\newblock Next-generation deep learning based on simulators and synthetic data.
\newblock \emph{Trends in Cognitive Sciences}, 26\penalty0 (2):\penalty0 174--187, February 2022.
\newblock ISSN 13646613.
\newblock \doi{10.1016/j.tics.2021.11.008}.
\newblock URL \url{https://linkinghub.elsevier.com/retrieve/pii/S136466132100293X}.

\bibitem[Du et~al.(2022)Du, Shao, Chai, Zhao, Diao, Gao, Yuan, Wang, Li, Zhang, Zhang, and Min]{du_synaptic_2022}
Yan Du, Wei Shao, Zheng Chai, Hanzhang Zhao, Qihui Diao, Yawei Gao, Xihui Yuan, Qiaoqiao Wang, Tao Li, Weidong Zhang, Jian~Fu Zhang, and Tai Min.
\newblock Synaptic 1/f noise injection for overfitting suppression in hardware neural networks.
\newblock \emph{Neuromorphic Computing and Engineering}, 2\penalty0 (3):\penalty0 034006, September 2022.
\newblock ISSN 2634-4386.
\newblock \doi{10.1088/2634-4386/ac6d05}.
\newblock URL \url{https://iopscience.iop.org/article/10.1088/2634-4386/ac6d05}.

\bibitem[El‐Atab et~al.(2020)El‐Atab, Canas, and Hussain]{elatab_pressuredriven_2020}
Nazek El‐Atab, Javier~Chavarrio Canas, and Muhammad~M. Hussain.
\newblock Pressure‐{Driven} {Two}‐{Input} {3D} {Microfluidic} {Logic} {Gates}.
\newblock \emph{Advanced Science}, 7\penalty0 (2):\penalty0 1903027, January 2020.
\newblock ISSN 2198-3844, 2198-3844.
\newblock \doi{10.1002/advs.201903027}.
\newblock URL \url{https://onlinelibrary.wiley.com/doi/10.1002/advs.201903027}.

\bibitem[Emeigh et~al.(2025)Emeigh, Ramsey, and Ryu]{emeigh_side-view_2025}
Carson Emeigh, Thomas Ramsey, and Sangjin Ryu.
\newblock Side-{View} {Imaging}-{Based} {Analysis} of the {Balloon} {Inflation}–{Deflation} {Dynamics} of a {Microfluidic} {Cell} {Compressor}.
\newblock \emph{Journal of Fluids Engineering}, 147\penalty0 (7):\penalty0 071108, July 2025.
\newblock ISSN 0098-2202, 1528-901X.
\newblock \doi{10.1115/1.4068616}.
\newblock URL \url{https://asmedigitalcollection.asme.org/fluidsengineering/article/147/7/071108/1217194/Side-View-Imaging-Based-Analysis-of-the-Balloon}.

\bibitem[Fu et~al.(2024)Fu, Fu, Wang, and Yao]{fu_enabling_2024}
Tianda Fu, Shuai Fu, Siqi Wang, and Jun Yao.
\newblock Enabling reliable two-terminal memristor network by exploiting the dynamic reverse recovery in a diode selector.
\newblock \emph{Device}, 2\penalty0 (4):\penalty0 100329, April 2024.
\newblock ISSN 26669986.
\newblock \doi{10.1016/j.device.2024.100329}.
\newblock URL \url{https://linkinghub.elsevier.com/retrieve/pii/S2666998624001212}.

\bibitem[Furuta et~al.(2018)Furuta, Fujii, Nakajima, Tsunegi, Kubota, Suzuki, and Miwa]{furuta_macromagnetic_2018}
Taishi Furuta, Keisuke Fujii, Kohei Nakajima, Sumito Tsunegi, Hitoshi Kubota, Yoshishige Suzuki, and Shinji Miwa.
\newblock Macromagnetic {Simulation} for {Reservoir} {Computing} {Utilizing} {Spin} {Dynamics} in {Magnetic} {Tunnel} {Junctions}.
\newblock \emph{Physical Review Applied}, 10\penalty0 (3):\penalty0 034063, September 2018.
\newblock ISSN 2331-7019.
\newblock \doi{10.1103/PhysRevApplied.10.034063}.
\newblock URL \url{https://link.aps.org/doi/10.1103/PhysRevApplied.10.034063}.

\bibitem[García-Beni et~al.(2023)García-Beni, Giorgi, Soriano, and Zambrini]{garcia-beni_scalable_2023}
Jorge García-Beni, Gian~Luca Giorgi, Miguel~C. Soriano, and Roberta Zambrini.
\newblock Scalable {Photonic} {Platform} for {Real}-{Time} {Quantum} {Reservoir} {Computing}.
\newblock \emph{Physical Review Applied}, 20\penalty0 (1):\penalty0 014051, July 2023.
\newblock ISSN 2331-7019.
\newblock \doi{10.1103/PhysRevApplied.20.014051}.
\newblock URL \url{https://link.aps.org/doi/10.1103/PhysRevApplied.20.014051}.

\bibitem[García-Hernández et~al.(2023)García-Hernández, Martínez-Martínez, Pazos-Solís, Aguado-Preciado, Dutt, Chávez-Ramírez, Korgel, Sharma, and Oza]{garcia-hernandez_optical_2023}
Luis García-Hernández, Eduardo Martínez-Martínez, Denni Pazos-Solís, Javier Aguado-Preciado, Ateet Dutt, Abraham Chávez-Ramírez, Brian Korgel, Ashutosh Sharma, and Goldie Oza.
\newblock Optical {Detection} of {Cancer} {Cells} {Using} {Lab}-on-a-{Chip}.
\newblock \emph{Biosensors}, 13\penalty0 (4):\penalty0 439, March 2023.
\newblock ISSN 2079-6374.
\newblock \doi{10.3390/bios13040439}.
\newblock URL \url{https://www.mdpi.com/2079-6374/13/4/439}.

\bibitem[Gauthier et~al.(2021)Gauthier, Bollt, Griffith, and Barbosa]{gauthier_next_2021}
Daniel~J. Gauthier, Erik Bollt, Aaron Griffith, and Wendson A.~S. Barbosa.
\newblock Next generation reservoir computing.
\newblock \emph{Nature Communications}, 12\penalty0 (1):\penalty0 5564, September 2021.
\newblock ISSN 2041-1723.
\newblock \doi{10.1038/s41467-021-25801-2}.
\newblock URL \url{https://www.nature.com/articles/s41467-021-25801-2}.

\bibitem[Goldt et~al.(2020)Goldt, Loureiro, Reeves, Krzakala, Mézard, and Zdeborová]{goldt_gaussian_2020}
Sebastian Goldt, Bruno Loureiro, Galen Reeves, Florent Krzakala, Marc Mézard, and Lenka Zdeborová.
\newblock The {Gaussian} equivalence of generative models for learning with shallow neural networks.
\newblock 2020.
\newblock \doi{10.48550/ARXIV.2006.14709}.
\newblock URL \url{https://arxiv.org/abs/2006.14709}.
\newblock Publisher: arXiv Version Number: 3.

\bibitem[Gómez et~al.(2025)Gómez, Hofmann, Debus, Başaran, Lotter, Khanzadeh, Angerbauer, Unluturk, Abadal, Haselmayr, Fitzek, Schober, and Dressler]{gomez_communicating_2025}
Jorge~Torres Gómez, Pit Hofmann, Lisa~Y. Debus, Osman~Tugay Başaran, Sebastian Lotter, Roya Khanzadeh, Stefan Angerbauer, Bige~Deniz Unluturk, Sergi Abadal, Werner Haselmayr, Frank H.~P. Fitzek, Robert Schober, and Falko Dressler.
\newblock Communicating {Smartly} in {Molecular} {Communication} {Environments}: {Neural} {Networks} in the {Internet} of {Bio}-{Nano} {Things}, 2025.
\newblock URL \url{https://arxiv.org/abs/2506.20589}.
\newblock Version Number: 3.

\bibitem[Hassija et~al.(2024)Hassija, Chamola, Mahapatra, Singal, Goel, Huang, Scardapane, Spinelli, Mahmud, and Hussain]{hassija_interpreting_2024}
Vikas Hassija, Vinay Chamola, Atmesh Mahapatra, Abhinandan Singal, Divyansh Goel, Kaizhu Huang, Simone Scardapane, Indro Spinelli, Mufti Mahmud, and Amir Hussain.
\newblock Interpreting {Black}-{Box} {Models}: {A} {Review} on {Explainable} {Artificial} {Intelligence}.
\newblock \emph{Cognitive Computation}, 16\penalty0 (1):\penalty0 45--74, January 2024.
\newblock ISSN 1866-9956, 1866-9964.
\newblock \doi{10.1007/s12559-023-10179-8}.
\newblock URL \url{https://link.springer.com/10.1007/s12559-023-10179-8}.

\bibitem[Hibat-Allah et~al.(2020)Hibat-Allah, Ganahl, Hayward, Melko, and Carrasquilla]{hibat-allah_recurrent_2020}
Mohamed Hibat-Allah, Martin Ganahl, Lauren~E. Hayward, Roger~G. Melko, and Juan Carrasquilla.
\newblock Recurrent neural network wave functions.
\newblock \emph{Physical Review Research}, 2\penalty0 (2):\penalty0 023358, June 2020.
\newblock ISSN 2643-1564.
\newblock \doi{10.1103/PhysRevResearch.2.023358}.
\newblock URL \url{https://link.aps.org/doi/10.1103/PhysRevResearch.2.023358}.

\bibitem[Jacobsen(2023)]{jacobsen_machine_2023}
Benjamin~N Jacobsen.
\newblock Machine learning and the politics of synthetic data.
\newblock \emph{Big Data \& Society}, 10\penalty0 (1):\penalty0 20539517221145372, January 2023.
\newblock ISSN 2053-9517, 2053-9517.
\newblock \doi{10.1177/20539517221145372}.
\newblock URL \url{https://journals.sagepub.com/doi/10.1177/20539517221145372}.

\bibitem[Jiang et~al.(2025)Jiang, Shi, Wang, Du, Wang, Lin, Li, Zhang, He, Sokolovskij, He, Li, Wang, Chen, Wang, Yu, and Wang]{jiang_-sensor_2025}
Yang Jiang, Shuhui Shi, Shaocong Wang, Fangzhou Du, Peiran Wang, Ning Lin, Wennao Li, Yi~Zhang, Leiwei He, Robert Sokolovskij, Jiaqi He, Mujun Li, Dingchen Wang, Xi~Chen, Qing Wang, Hongyu Yu, and Zhongrui Wang.
\newblock In-sensor reservoir computing for gas pattern recognition using {Pt}-{AlGaN}/{GaN} {HEMTs}.
\newblock \emph{Device}, 3\penalty0 (1):\penalty0 100550, January 2025.
\newblock ISSN 26669986.
\newblock \doi{10.1016/j.device.2024.100550}.
\newblock URL \url{https://linkinghub.elsevier.com/retrieve/pii/S2666998624004721}.

\bibitem[Kopets et~al.(2022)Kopets, Tatiana, Rybin, Dautov, Karimov, and Karimov]{kopets_simulation_2022}
Ekaterina Kopets, Shchetinina Tatiana, Vyacheslav Rybin, Albert Dautov, Timur Karimov, and Artur Karimov.
\newblock Simulation of a {Small}-{Scale} {Chemical} {Reservoir} {Computer} for {Pattern} {Recognition}.
\newblock In \emph{2022 11th {Mediterranean} {Conference} on {Embedded} {Computing} ({MECO})}, pages 1--4, Budva, Montenegro, June 2022. IEEE.
\newblock ISBN 978-1-6654-6828-2.
\newblock \doi{10.1109/MECO55406.2022.9797166}.
\newblock URL \url{https://ieeexplore.ieee.org/document/9797166/}.

\bibitem[Kuscu and Akan(2016)]{kuscu_modeling_2016}
Murat Kuscu and Ozgur~B. Akan.
\newblock Modeling and {Analysis} of {SiNW} {FET}-{Based} {Molecular} {Communication} {Receiver}.
\newblock \emph{IEEE Transactions on Communications}, 64\penalty0 (9):\penalty0 3708--3721, September 2016.
\newblock ISSN 0090-6778.
\newblock \doi{10.1109/TCOMM.2016.2589935}.
\newblock URL \url{http://ieeexplore.ieee.org/document/7508935/}.

\bibitem[Kuscu et~al.(2021)Kuscu, Ramezani, Dinc, Akhavan, and Akan]{kuscu_fabrication_2021}
Murat Kuscu, Hamideh Ramezani, Ergin Dinc, Shahab Akhavan, and Ozgur~B. Akan.
\newblock Fabrication and microfluidic analysis of graphene-based molecular communication receiver for {Internet} of {Nano} {Things} ({IoNT}).
\newblock \emph{Scientific Reports}, 11\penalty0 (1):\penalty0 19600, October 2021.
\newblock ISSN 2045-2322.
\newblock \doi{10.1038/s41598-021-98609-1}.
\newblock URL \url{https://www.nature.com/articles/s41598-021-98609-1}.

\bibitem[Köster et~al.(2023)Köster, Patel, Wikner, Jaurigue, and Lüdge]{koster_data-informed_2023}
Felix Köster, Dhruvit Patel, Alexander Wikner, Lina Jaurigue, and Kathy Lüdge.
\newblock Data-informed reservoir computing for efficient time-series prediction.
\newblock \emph{Chaos: An Interdisciplinary Journal of Nonlinear Science}, 33\penalty0 (7):\penalty0 073109, July 2023.
\newblock ISSN 1054-1500, 1089-7682.
\newblock \doi{10.1063/5.0152311}.
\newblock URL \url{https://pubs.aip.org/cha/article/33/7/073109/2901134/Data-informed-reservoir-computing-for-efficient}.

\bibitem[Leung et~al.(2022)Leung, De~Haan, Ronaldson-Bouchard, Kim, Ko, Rho, Chen, Habibovic, Jeon, Takayama, Shuler, Vunjak-Novakovic, Frey, Verpoorte, and Toh]{leung_guide_2022}
Chak~Ming Leung, Pim De~Haan, Kacey Ronaldson-Bouchard, Ge-Ah Kim, Jihoon Ko, Hoon~Suk Rho, Zhu Chen, Pamela Habibovic, Noo~Li Jeon, Shuichi Takayama, Michael~L. Shuler, Gordana Vunjak-Novakovic, Olivier Frey, Elisabeth Verpoorte, and Yi-Chin Toh.
\newblock A guide to the organ-on-a-chip.
\newblock \emph{Nature Reviews Methods Primers}, 2\penalty0 (1):\penalty0 33, May 2022.
\newblock ISSN 2662-8449.
\newblock \doi{10.1038/s43586-022-00118-6}.
\newblock URL \url{https://www.nature.com/articles/s43586-022-00118-6}.

\bibitem[Liao et~al.(2023)Liao, Yamahara, Terao, Ma, Seki, and Tabata]{liao_short-term_2023}
Zhiqiang Liao, Hiroyasu Yamahara, Kenyu Terao, Kaijie Ma, Munetoshi Seki, and Hitoshi Tabata.
\newblock Short-term memory capacity analysis of {Lu3Fe4Co0}.{5Si0}.{5O12}-based spin cluster glass towards reservoir computing.
\newblock \emph{Scientific Reports}, 13\penalty0 (1):\penalty0 5260, March 2023.
\newblock ISSN 2045-2322.
\newblock \doi{10.1038/s41598-023-32084-8}.
\newblock URL \url{https://www.nature.com/articles/s41598-023-32084-8}.

\bibitem[Loyola-Gonzalez(2019)]{loyola-gonzalez_black-box_2019}
Octavio Loyola-Gonzalez.
\newblock Black-{Box} vs. {White}-{Box}: {Understanding} {Their} {Advantages} and {Weaknesses} {From} a {Practical} {Point} of {View}.
\newblock \emph{IEEE Access}, 7:\penalty0 154096--154113, 2019.
\newblock ISSN 2169-3536.
\newblock \doi{10.1109/ACCESS.2019.2949286}.
\newblock URL \url{https://ieeexplore.ieee.org/document/8882211/}.

\bibitem[Lu et~al.(2023)Lu, Chen, Zhang, Shen, Wang, Wang, van Rechem, Fu, and Wei]{lu_machine_2023}
Yingzhou Lu, Lulu Chen, Yuanyuan Zhang, Minjie Shen, Huazheng Wang, Xiao Wang, Capucine van Rechem, Tianfan Fu, and Wenqi Wei.
\newblock Machine {Learning} for {Synthetic} {Data} {Generation}: {A} {Review}, 2023.
\newblock URL \url{https://arxiv.org/abs/2302.04062}.
\newblock Version Number: 10.

\bibitem[Maass et~al.(2002)Maass, Natschläger, and Markram]{maass_real-time_2002}
Wolfgang Maass, Thomas Natschläger, and Henry Markram.
\newblock Real-{Time} {Computing} {Without} {Stable} {States}: {A} {New} {Framework} for {Neural} {Computation} {Based} on {Perturbations}.
\newblock \emph{Neural Computation}, 14\penalty0 (11):\penalty0 2531--2560, November 2002.
\newblock ISSN 0899-7667, 1530-888X.
\newblock \doi{10.1162/089976602760407955}.
\newblock URL \url{https://direct.mit.edu/neco/article/14/11/2531-2560/6650}.

\bibitem[Matsumura et~al.(2025)Matsumura, Honda, Kikuchi, Mizuno, Hara, Kondo, Nakamura, Watanabe, Hayakawa, Nakajima, and Takei]{matsumura_real-time_2025}
Guren Matsumura, Satoko Honda, Takamasa Kikuchi, Yuuki Mizuno, Hyuga Hara, Yoshiki Kondo, Haruki Nakamura, Shin Watanabe, Kiyoshi Hayakawa, Kohei Nakajima, and Kuniharu Takei.
\newblock Real-time personal healthcare data analysis using edge computing for multimodal wearable sensors.
\newblock \emph{Device}, 3\penalty0 (2):\penalty0 100597, February 2025.
\newblock ISSN 26669986.
\newblock \doi{10.1016/j.device.2024.100597}.
\newblock URL \url{https://linkinghub.elsevier.com/retrieve/pii/S266699862400543X}.

\bibitem[Miali et~al.(2019)Miali, Colasuonno, Surdo, Palomba, Pereira, Rondanina, Diaspro, Pascazio, and Decuzzi]{miali_leaf-inspired_2019}
Marco~E. Miali, Marianna Colasuonno, Salvatore Surdo, Roberto Palomba, Rui Pereira, Eliana Rondanina, Alberto Diaspro, Giuseppe Pascazio, and Paolo Decuzzi.
\newblock Leaf-{Inspired} {Authentically} {Complex} {Microvascular} {Networks} for {Deciphering} {Biological} {Transport} {Process}.
\newblock \emph{ACS Applied Materials \& Interfaces}, 11\penalty0 (35):\penalty0 31627--31637, September 2019.
\newblock ISSN 1944-8244, 1944-8252.
\newblock \doi{10.1021/acsami.9b09453}.
\newblock URL \url{https://pubs.acs.org/doi/10.1021/acsami.9b09453}.

\bibitem[Mienye et~al.(2024)Mienye, Swart, and Obaido]{mienye_recurrent_2024}
Ibomoiye~Domor Mienye, Theo~G. Swart, and George Obaido.
\newblock Recurrent {Neural} {Networks}: {A} {Comprehensive} {Review} of {Architectures}, {Variants}, and {Applications}.
\newblock \emph{Information}, 15\penalty0 (9):\penalty0 517, August 2024.
\newblock ISSN 2078-2489.
\newblock \doi{10.3390/info15090517}.
\newblock URL \url{https://www.mdpi.com/2078-2489/15/9/517}.

\bibitem[Misba et~al.(2023)Misba, Mavikumbure, Rajib, Marino, Cobilean, Manic, and Atulasimha]{misba_spintronic_2023}
Walid~Al Misba, Harindra~S. Mavikumbure, Md~Mahadi Rajib, Daniel~L. Marino, Victor Cobilean, Milos Manic, and Jayasimha Atulasimha.
\newblock Spintronic {Physical} {Reservoir} for {Autonomous} {Prediction} and {Long}-{Term} {Household} {Energy} {Load} {Forecasting}.
\newblock \emph{IEEE Access}, 11:\penalty0 124725--124737, 2023.
\newblock ISSN 2169-3536.
\newblock \doi{10.1109/ACCESS.2023.3326414}.
\newblock URL \url{https://ieeexplore.ieee.org/document/10311403/}.

\bibitem[Nakajima et~al.(2022)Nakajima, Inoue, Tanaka, Kuniyoshi, Hashimoto, and Nakajima]{nakajima_physical_2022}
Mitsumasa Nakajima, Katsuma Inoue, Kenji Tanaka, Yasuo Kuniyoshi, Toshikazu Hashimoto, and Kohei Nakajima.
\newblock Physical deep learning with biologically inspired training method: gradient-free approach for physical hardware.
\newblock \emph{Nature Communications}, 13\penalty0 (1):\penalty0 7847, December 2022.
\newblock ISSN 2041-1723.
\newblock \doi{10.1038/s41467-022-35216-2}.
\newblock URL \url{https://www.nature.com/articles/s41467-022-35216-2}.

\bibitem[Nikolić et~al.(2023)Nikolić, Echlin, Aguilar, and Shmulevich]{nikolic_computational_2023}
Vladimir Nikolić, Moriah Echlin, Boris Aguilar, and Ilya Shmulevich.
\newblock Computational capabilities of a multicellular reservoir computing system.
\newblock \emph{PLOS ONE}, 18\penalty0 (4):\penalty0 e0282122, April 2023.
\newblock ISSN 1932-6203.
\newblock \doi{10.1371/journal.pone.0282122}.
\newblock URL \url{https://dx.plos.org/10.1371/journal.pone.0282122}.

\bibitem[Raab et~al.(2022)Raab, Brems, Beneke, Dohi, Rothörl, Kammerbauer, Mentink, and Kläui]{raab_brownian_2022}
Klaus Raab, Maarten~A. Brems, Grischa Beneke, Takaaki Dohi, Jan Rothörl, Fabian Kammerbauer, Johan~H. Mentink, and Mathias Kläui.
\newblock Brownian reservoir computing realized using geometrically confined skyrmion dynamics.
\newblock \emph{Nature Communications}, 13\penalty0 (1):\penalty0 6982, November 2022.
\newblock ISSN 2041-1723.
\newblock \doi{10.1038/s41467-022-34309-2}.
\newblock URL \url{https://www.nature.com/articles/s41467-022-34309-2}.

\bibitem[Rajotte et~al.(2022)Rajotte, Bergen, Buckeridge, El~Emam, Ng, and Strome]{rajotte_synthetic_2022}
Jean-Francois Rajotte, Robert Bergen, David~L. Buckeridge, Khaled El~Emam, Raymond Ng, and Elissa Strome.
\newblock Synthetic data as an enabler for machine learning applications in medicine.
\newblock \emph{iScience}, 25\penalty0 (11):\penalty0 105331, November 2022.
\newblock ISSN 25890042.
\newblock \doi{10.1016/j.isci.2022.105331}.
\newblock URL \url{https://linkinghub.elsevier.com/retrieve/pii/S2589004222016030}.

\bibitem[Ryu et~al.(2025)Ryu, Zhang, Palmon, Salcedo, Pass, and Socha]{ryu_insect_2025}
Sangjin Ryu, Haipeng Zhang, Tomer Palmon, Mary~K. Salcedo, Günther Pass, and John~J. Socha.
\newblock Insect wing circulation: transient perfusion through a microfluidic dragonfly forewing model.
\newblock \emph{Lab on a Chip}, 25\penalty0 (15):\penalty0 3718--3729, 2025.
\newblock ISSN 1473-0197, 1473-0189.
\newblock \doi{10.1039/D4LC00714J}.
\newblock URL \url{https://xlink.rsc.org/?DOI=D4LC00714J}.

\bibitem[Singh et~al.(2023)Singh, Tintelott, Moussavi, Ingebrandt, Leupers, Vu, Merchant, and Pachauri]{singh_logic_2023}
Animesh~Pratap Singh, Marcel Tintelott, Elmira Moussavi, Sven Ingebrandt, Rainer Leupers, Xuan-Thang Vu, Farhad Merchant, and Vivek Pachauri.
\newblock Logic operations in fluidics as foundation for embedded biohybrid computation.
\newblock \emph{Device}, 1\penalty0 (6):\penalty0 100220, December 2023.
\newblock ISSN 26669986.
\newblock \doi{10.1016/j.device.2023.100220}.
\newblock URL \url{https://linkinghub.elsevier.com/retrieve/pii/S2666998623003605}.

\bibitem[So et~al.(2023)So, Lee, and Kim]{so_short-term_2023}
Hyojin So, Jung-Kyu Lee, and Sungjun Kim.
\newblock Short-term memory characteristics in n-type-{ZnO}/p-type-{NiO} heterojunction synaptic devices for reservoir computing.
\newblock \emph{Applied Surface Science}, 625:\penalty0 157153, July 2023.
\newblock ISSN 01694332.
\newblock \doi{10.1016/j.apsusc.2023.157153}.
\newblock URL \url{https://linkinghub.elsevier.com/retrieve/pii/S0169433223008310}.

\bibitem[Sumi et~al.(2023)Sumi, Yamamoto, Katori, Ito, Moriya, Konno, Sato, and Hirano-Iwata]{sumi_biological_2023}
Takuma Sumi, Hideaki Yamamoto, Yuichi Katori, Koki Ito, Satoshi Moriya, Tomohiro Konno, Shigeo Sato, and Ayumi Hirano-Iwata.
\newblock Biological neurons act as generalization filters in reservoir computing.
\newblock \emph{Proceedings of the National Academy of Sciences}, 120\penalty0 (25):\penalty0 e2217008120, June 2023.
\newblock ISSN 0027-8424, 1091-6490.
\newblock \doi{10.1073/pnas.2217008120}.
\newblock URL \url{https://pnas.org/doi/10.1073/pnas.2217008120}.

\bibitem[Sun et~al.(2021)Sun, Wang, Jiang, Kim, Joo, Zheng, Lee, Yu, Kong, and Yang]{sun_-sensor_2021}
Linfeng Sun, Zhongrui Wang, Jinbao Jiang, Yeji Kim, Bomin Joo, Shoujun Zheng, Seungyeon Lee, Woo~Jong Yu, Bai-Sun Kong, and Heejun Yang.
\newblock In-sensor reservoir computing for language learning via two-dimensional memristors.
\newblock \emph{Science Advances}, 7\penalty0 (20):\penalty0 eabg1455, May 2021.
\newblock ISSN 2375-2548.
\newblock \doi{10.1126/sciadv.abg1455}.
\newblock URL \url{https://www.science.org/doi/10.1126/sciadv.abg1455}.

\bibitem[Temiz et~al.(2015)Temiz, Lovchik, Kaigala, and Delamarche]{temiz_lab---chip_2015}
Yuksel Temiz, Robert~D. Lovchik, Govind~V. Kaigala, and Emmanuel Delamarche.
\newblock Lab-on-a-chip devices: {How} to close and plug the lab?
\newblock \emph{Microelectronic Engineering}, 132:\penalty0 156--175, January 2015.
\newblock ISSN 01679317.
\newblock \doi{10.1016/j.mee.2014.10.013}.
\newblock URL \url{https://linkinghub.elsevier.com/retrieve/pii/S0167931714004456}.

\bibitem[Tsakalos et~al.(2022)Tsakalos, Sirakoulis, Adamatzky, and Smith]{tsakalos_protein_2022}
Karolos-Alexandros Tsakalos, Georgios~Ch. Sirakoulis, Andrew Adamatzky, and Jim Smith.
\newblock Protein {Structured} {Reservoir} {Computing} for {Spike}-{Based} {Pattern} {Recognition}.
\newblock \emph{IEEE Transactions on Parallel and Distributed Systems}, 33\penalty0 (2):\penalty0 322--331, February 2022.
\newblock ISSN 1045-9219, 1558-2183, 2161-9883.
\newblock \doi{10.1109/TPDS.2021.3068826}.
\newblock URL \url{https://ieeexplore.ieee.org/document/9387584/}.

\bibitem[Uzun et~al.(2025)Uzun, Ikiz, and Kuscu]{uzun_molecular_2025}
Mustafa Uzun, Kaan~Burak Ikiz, and Murat Kuscu.
\newblock Molecular {Communication} {Channel} as a {Physical} {Reservoir} {Computer}, 2025.
\newblock URL \url{https://arxiv.org/abs/2504.17022}.
\newblock Version Number: 1.

\bibitem[Xu et~al.(2019)Xu, Skoularidou, Cuesta-Infante, and Veeramachaneni]{xu_modeling_2019}
Lei Xu, Maria Skoularidou, Alfredo Cuesta-Infante, and Kalyan Veeramachaneni.
\newblock Modeling {Tabular} data using {Conditional} {GAN}, 2019.
\newblock URL \url{https://arxiv.org/abs/1907.00503}.
\newblock Version Number: 2.

\bibitem[Xue and Salim(2024)]{xue_promptcast_2024}
Hao Xue and Flora~D. Salim.
\newblock {PromptCast}: {A} {New} {Prompt}-{Based} {Learning} {Paradigm} for {Time} {Series} {Forecasting}.
\newblock \emph{IEEE Transactions on Knowledge and Data Engineering}, 36\penalty0 (11):\penalty0 6851--6864, November 2024.
\newblock ISSN 1041-4347, 1558-2191, 2326-3865.
\newblock \doi{10.1109/TKDE.2023.3342137}.
\newblock URL \url{https://ieeexplore.ieee.org/document/10356715/}.

\bibitem[Zhang and Akan(2025)]{zhang_ion_2025}
Shaojie Zhang and Ozgur~B. Akan.
\newblock Ion {Transmitter} for {Molecular} {Communication}, 2025.
\newblock URL \url{https://arxiv.org/abs/2501.10392}.
\newblock Version Number: 1.

\bibitem[Zhang et~al.(2020)Zhang, Liu, and Suen]{zhang_towards_2020}
Xu-Yao Zhang, Cheng-Lin Liu, and Ching~Y. Suen.
\newblock Towards {Robust} {Pattern} {Recognition}: {A} {Review}.
\newblock \emph{Proceedings of the IEEE}, 108\penalty0 (6):\penalty0 894--922, June 2020.
\newblock ISSN 0018-9219, 1558-2256.
\newblock \doi{10.1109/JPROC.2020.2989782}.
\newblock URL \url{https://ieeexplore.ieee.org/document/9103349/}.

\bibitem[Zhong et~al.(2021)Zhong, Tang, Li, Gao, Qian, and Wu]{zhong_dynamic_2021}
Yanan Zhong, Jianshi Tang, Xinyi Li, Bin Gao, He~Qian, and Huaqiang Wu.
\newblock Dynamic memristor-based reservoir computing for high-efficiency temporal signal processing.
\newblock \emph{Nature Communications}, 12\penalty0 (1):\penalty0 408, January 2021.
\newblock ISSN 2041-1723.
\newblock \doi{10.1038/s41467-020-20692-1}.
\newblock URL \url{https://www.nature.com/articles/s41467-020-20692-1}.

\bibitem[Zhong et~al.(2022)Zhong, Tang, Li, Liang, Liu, Li, Xi, Yao, Hao, Gao, Qian, and Wu]{zhong_memristor-based_2022}
Yanan Zhong, Jianshi Tang, Xinyi Li, Xiangpeng Liang, Zhengwu Liu, Yijun Li, Yue Xi, Peng Yao, Zhenqi Hao, Bin Gao, He~Qian, and Huaqiang Wu.
\newblock A memristor-based analogue reservoir computing system for real-time and power-efficient signal processing.
\newblock \emph{Nature Electronics}, 5\penalty0 (10):\penalty0 672--681, September 2022.
\newblock ISSN 2520-1131.
\newblock \doi{10.1038/s41928-022-00838-3}.
\newblock URL \url{https://www.nature.com/articles/s41928-022-00838-3}.

\end{thebibliography}


\begin{thebibliography}{5}
\providecommand{\natexlab}[1]{#1}
\providecommand{\url}[1]{\texttt{#1}}
\expandafter\ifx\csname urlstyle\endcsname\relax
  \providecommand{\doi}[1]{doi: #1}\else
  \providecommand{\doi}{doi: \begingroup \urlstyle{rm}\Url}\fi

\bibitem[Becerra-Suarez et~al.(2025)Becerra-Suarez, Alvarez-Vasquez, and Forero]{becerra-suarez_improvement_2025}
Fray~L. Becerra-Suarez, Halyn Alvarez-Vasquez, and Manuel~G. Forero.
\newblock Improvement of {Bank} {Fraud} {Detection} {Through} {Synthetic} {Data} {Generation} with {Gaussian} {Noise}.
\newblock \emph{Technologies}, 13\penalty0 (4):\penalty0 141, April 2025.
\newblock ISSN 2227-7080.
\newblock \doi{10.3390/technologies13040141}.
\newblock URL \url{https://www.mdpi.com/2227-7080/13/4/141}.

\bibitem[Du et~al.(2022)Du, Shao, Chai, Zhao, Diao, Gao, Yuan, Wang, Li, Zhang, Zhang, and Min]{du_synaptic_2022}
Yan Du, Wei Shao, Zheng Chai, Hanzhang Zhao, Qihui Diao, Yawei Gao, Xihui Yuan, Qiaoqiao Wang, Tao Li, Weidong Zhang, Jian~Fu Zhang, and Tai Min.
\newblock Synaptic 1/f noise injection for overfitting suppression in hardware neural networks.
\newblock \emph{Neuromorphic Computing and Engineering}, 2\penalty0 (3):\penalty0 034006, September 2022.
\newblock ISSN 2634-4386.
\newblock \doi{10.1088/2634-4386/ac6d05}.
\newblock URL \url{https://iopscience.iop.org/article/10.1088/2634-4386/ac6d05}.

\bibitem[Lutz et~al.(2021)Lutz, Sauvage, and Dischler]{lutz_cyclostationary_2021}
Nicolas Lutz, Basile Sauvage, and Jean‐Michel Dischler.
\newblock Cyclostationary {Gaussian} noise: theory and synthesis.
\newblock \emph{Computer Graphics Forum}, 40\penalty0 (2):\penalty0 239--250, May 2021.
\newblock ISSN 0167-7055, 1467-8659.
\newblock \doi{10.1111/cgf.142629}.
\newblock URL \url{https://onlinelibrary.wiley.com/doi/10.1111/cgf.142629}.

\bibitem[Patki et~al.(2016)Patki, Wedge, and Veeramachaneni]{patki_synthetic_2016}
Neha Patki, Roy Wedge, and Kalyan Veeramachaneni.
\newblock The {Synthetic} {Data} {Vault}.
\newblock In \emph{2016 {IEEE} {International} {Conference} on {Data} {Science} and {Advanced} {Analytics} ({DSAA})}, pages 399--410, Montreal, QC, Canada, October 2016. IEEE.
\newblock ISBN 978-1-5090-5206-6.
\newblock \doi{10.1109/DSAA.2016.49}.
\newblock URL \url{http://ieeexplore.ieee.org/document/7796926/}.

\bibitem[Xu et~al.(2019)Xu, Skoularidou, Cuesta-Infante, and Veeramachaneni]{xu_modeling_2019}
Lei Xu, Maria Skoularidou, Alfredo Cuesta-Infante, and Kalyan Veeramachaneni.
\newblock Modeling {Tabular} data using {Conditional} {GAN}, 2019.
\newblock URL \url{https://arxiv.org/abs/1907.00503}.
\newblock Version Number: 2.

\end{thebibliography}

\section*{RESOURCE AVAILABILITY}

\subsection*{Lead contact}

Requests for additional information and resources should be directed to the lead contact, Jacob Clouse (jclouse2@huskers.unl.edu).

\subsection*{Materials availability}

This study did not generate new unique reagents or materials.

\subsection*{Data and code availability}

RGB detection code as well as video examples of trials are located at: \\https://github.com/jclouse20/Microfluidic-System-for-Reservoir-Computing.

\section*{ACKNOWLEDGMENTS}

This publication has emanated from research conducted with the
financial support of University of Nebraska-Lincoln start-up grant as well as the Nebraska Research Initiative.

\section*{AUTHOR CONTRIBUTIONS}

J.C. and T.R. gathered experimental results.
S.S., S.B., and S.R. designed the theoretical framework of the study. 
T.R. and S.R. designed and created the microfluidic chip used in the study. 
N.K. created the dye detection code used to process video into RGB signals.
J.C. developed the readout layer and performed system analysis.
All the authors reviewed the final manuscript.

\section*{DECLARATION OF INTERESTS}

The authors declare no competing interests.

\section*{DECLARATION OF GENERATIVE AI AND AI-ASSISTED TECHNOLOGIES}

No significant AI usage.

\section*{SUPPLEMENTAL INFORMATION INDEX}

Document S1. Figures S1–S13, Notes S1-S9



\end{document}


\nolinenumbers

\author[1]{Jacob Clouse}
\author[2]{Thomas Ramsey}
\author[1] {Samitha Somathilaka}
\author[1] {Nicholas Kleinsasser}
\author[2] {Sangjin Ryu}
\author[1] {Sasitharan Balasubramaniam}

\affil[1]{School of Computing, University of Nebraska Lincoln, Lincoln Nebraska, USA}
\affil[2]{Department of Mechanical and Materials Engineering, University of Nebraska Lincoln, Lincoln Nebraska, USA}

\affil[*]{Correspondence: jclouse2@huskers.unl.edu, tramsey3@huskers.unl.edu, ssomathilaka2@unl.edu, nkleinsasser2@huskers.unl.edu, sryu2@unl.edu, sasi@unl.edu}

\maketitle

\section{Supplemental Notes}

The supplemental notes expand on the procedures and tests performed by providing key supporting details. We first explain how the size of the readout layer is selected across different experimental configurations and describe how repeated tests are conducted during readout layer training. In addition, we outline the methods used to generate synthetic records based on real records collected from experimental tests. We then perform a closer analysis on how quantization intervals of $2$, $5$, and $10$ influence mutual information (MI) within the system. This is followed by an analysis of how different pattern selections affect classification, first through a similarity comparison before the patterns enter the reservoir and then through a mean absolute difference (MAD) comparison after passing through the reservoir. The notes conclude with a discussion on the process of injecting the microfluidic chip with water to remove any air bubbles.
\newline

\textbf{Note S1. Defining the size of the Readout Layer} 

This note outlines the key factors that influence the number of input features and output nodes of the readout layer. The number of input features in the readout layer is determined by the chosen quantization intervals and the number of selected reservoir outputs. Quantization involves dividing and averaging each of the nine reservoir outputs according to a chosen quantization interval. In this research, we use quantization intervals of $1$, $2$, $5$, and $10$. The number of input features in the readout layer corresponds to the product of the quantization level and the number of reservoir outputs: $\textit{Number of input features} = Q \times O$ where \( Q \) is the number of quantization intervals and \( O \) is the number of reservoir outputs. For example, with \( Q = 5 \) and \( O = 9 \), we obtain \( 5 \times 9 = 45 \) input features. The number of output nodes in the readout layer is determined by the number of patterns being classified. In our case, eight patterns result in eight output nodes. 
In the current reservoir model, the largest readout layer occurs when all nine reservoir outputs are utilized, and each output is divided into $10$ quantization intervals. This produces $90$ distinct input features. When these are fully connected to eight output nodes for classification, the readout layer contains $720$ edges.
\newline

\renewcommand{\thefigure}{S\arabic{figure}}

\begin{figure}[H]
    \centering
    \includegraphics[width=1\textwidth ]{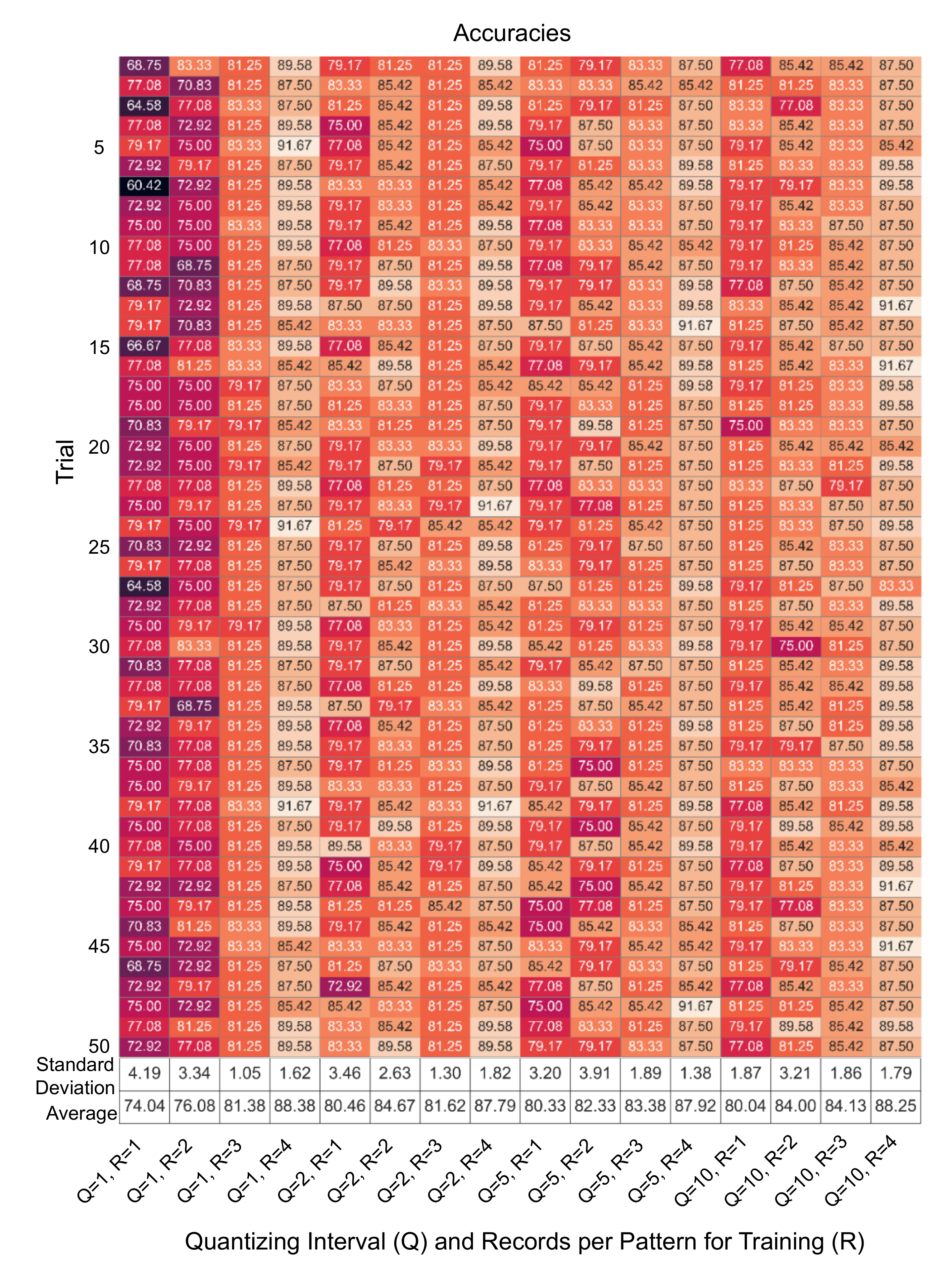}
    \label{fig:boxplot} 
\end{figure} 
\pagebreak
\begin{figure}[t!] 
    \caption{(Previous page.) This heatmap is an expansion of the results presented in Figure 4(G) in the main text. Figure 4(G) uses the average of $50$ uniquely trained models in order to minimize the effects of the random initial weights on network edges on results. This figure shows a heatmap for each quantization interval compared to the number of records per pattern used to train the readout layer over 50 separate models. Q represents the quantizing interval while R represents the variations per pattern used. It also shows the average and standard deviation from the average for each column in the heatmap.}
    \label{fig:boxplot} 
\end{figure}

\textbf{Note S2. Accuracy when Varying Quantizing Intervals and Records Per Pattern} 

Figure~\ref{fig:boxplot} illustrates the classification accuracy of over $50$ trained models for varying numbers of quantization intervals and real record. This shows the individual accuracies behind the average accuracy values presented in Figure 4(G). When performing experimental tests, we train multiple models to reduce randomness in the results and to verify the integrity of the system through repetition. A trend realized in the heatmap is that increasing both quantization intervals and the number of real records leads to higher and more consistent accuracies. Using four records per pattern results in an average accuracy of approximately $88\%$ across all quantization intervals, while using one record per pattern drops the accuracy across all quantization intervals to below $81\%$. Figure~\ref{fig:boxplot} also presents the standard deviation along with the average for each configuration. As the number of records per pattern used to train the models increases, the standard deviation of the average accuracy generally decreases. With a quantization interval of $1$, the use of a single record per pattern results in a standard deviation of $4.19\%$ centered around an average accuracy of $74.04\%$, while the use of four records per pattern reduces the standard deviation to $1.62\%$ with an average accuracy of $88.38\%$. This trend holds across most configurations, except when using a quantization interval of $10$. In that case, one record yields a standard deviation of $1.87\%$ centered on $80.04\%$, but two records increase it to $3.21\%$ with an average accuracy of $84\%$. This value still drops with higher record counts, with four records per pattern leading to a standard deviation of $1.79\%$ with an average accuracy of $88.25\%$. This strongly indicates that as the number of records per pattern increases, the standard deviation decreases and the accuracies become more consistent with less volatility. Based on these results, we use four records per pattern when training our models and generating synthetic data, as this number yields the best performance.

\begin{figure}[H]
\centering
\includegraphics[width=0.4\textwidth]{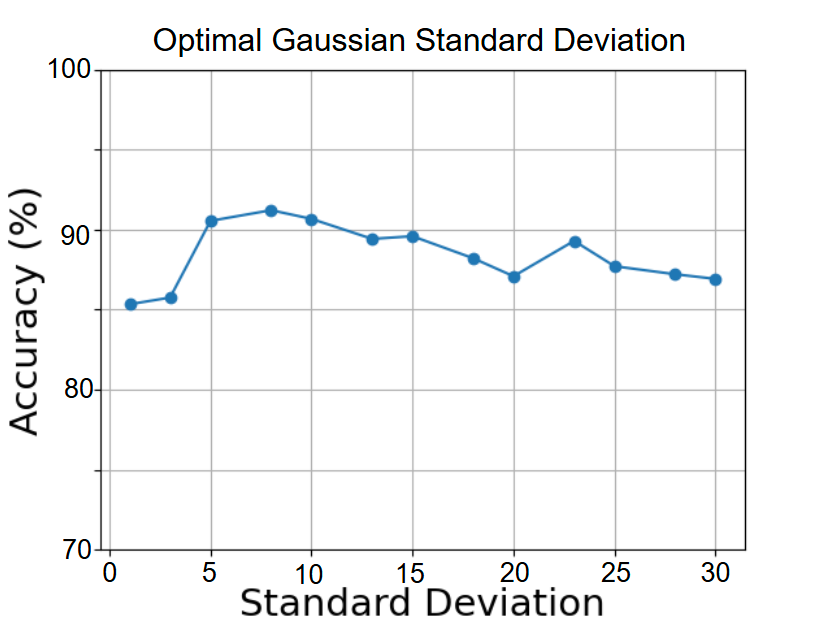}
\caption{Illustration comparing accuracy to \emph{Gaussian} standard deviation when finding the optimal \emph{Gaussian} standard deviation for data augmentation}
\label{fig:gaussian_standard_deviation} 
\end{figure}

\textbf{Note S3. Synthetic Data Generation}

To supplement the limited number of experimental records, we analyze synthetic data generation methods to expand the training set and improve the performance of our model. Our goal is to evaluate whether artificially created records can effectively mimic the variability of physical experiments, while preserving pattern-defining features. Our first approach to data synthesis involves applying \emph{Gaussian} noise to existing quantized reservoir outputs taken from the training dataset\cite{becerra-suarez_improvement_2025, lutz_cyclostationary_2021}. We generate this noise using a distribution centered on the quantized outputs collected from the microfluidic chip and apply it as a single value across each signal. This causes the entire signal to shift uniformly up or down replicating the natural behavior of our system as seen in Figure 3(B). An additional benefit of adding \emph{Gaussian} noise is that it introduces randomness to the dataset, which can help prevent overfitting\cite{du_synaptic_2022} during training. To determine the most effective standard deviation of the gaussian distribution for this method, we test multiple values ranging from $1$ to $30$, as shown in Figure~\ref{fig:gaussian_standard_deviation}. Analyzing the graph, we find that a lower standard deviation of $1$, which produces records close to the originals, does not improve classification performance. Once the standard deviation is increased to $8$, the system reaches a classification accuracy of around $90\%$ with $200$ total records. Further increasing the standard deviation past $8$ has little effect on accuracy; however, once the standard deviation exceeds $15$, classification accuracy drops below $90\%$. Based on these results, we use a standard deviation of $8$ for all subsequent tests involving \emph{Gaussian}-generated synthetic records.

When analyzing synthetic data generation, we also investigate \emph{Conditional Tabular GAN} (CTGAN)\cite{xu_modeling_2019}, a model provided by Synthetic Data Vault (SDV)\cite{patki_synthetic_2016}. Unlike the \emph{Gaussian} noise method, which perturbs existing quantized reservoir outputs, CTGAN is a generative adversarial network that learns the statistical distribution of each output to generate entirely new records. CTGAN is originally designed for tabular data, making it compatible with our quantized dataset. This approach enables the creation of more diverse and complex synthetic data. To generate data with CTGAN, we train a separate model for each of the $8$ patterns in our dataset. Each model requires at least two real records per pattern to begin training, and increasing the number of real records generally improves performance. However, we find that CTGAN-generated datasets consistently fail to exceed $90\%$ accuracy and never outperform the \emph{Gaussian} noise method. This limitation was primarily attributed to the small size of the training data used to generate synthetic records, with only four real records per pattern available for training. While both CTGAN and \emph{Gaussian} noise approaches reduce the need to collect additional real-world data, we find that \emph{Gaussian} noise is the preferred method for our use. It is both faster and more computationally efficient. Although readout layer processing time is equivalent for both methods, CTGAN introduces additional complexity during data generation, as it requires training eight separate models, one for each pattern, which adds significant time and resource overhead. 

\begin{figure}[H]
\centering
\includegraphics[width=1\textwidth]{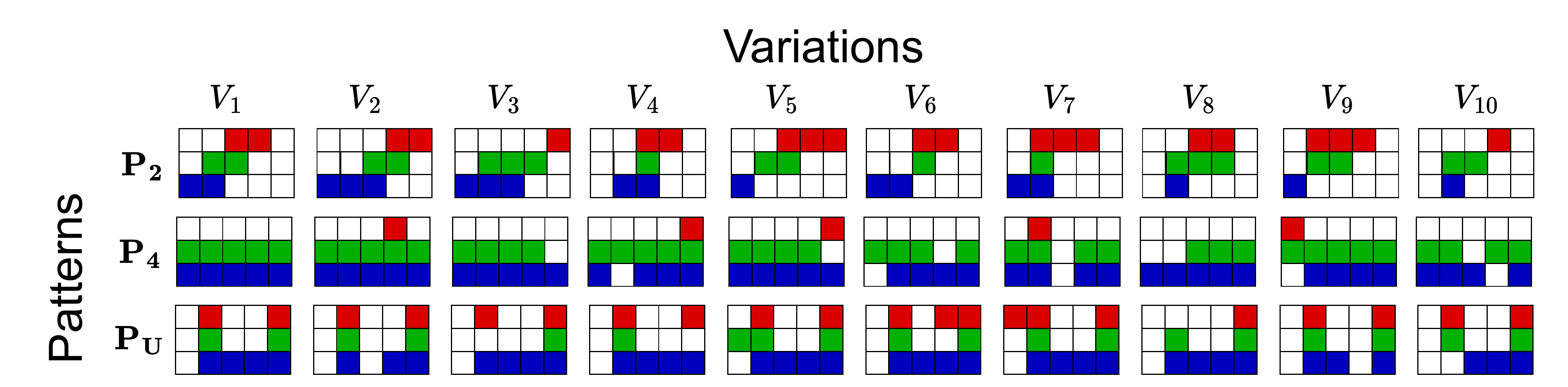}
\caption{
Patterns $\bf{P_2}$, $\bf{P_4}$, and $\bf{P_U}$ that are being used to calculate MI in Figures S4, S5, and S6. These patterns were selected to show one example of a diagonal shape, a constant signal, and a letter when calculating MI and how they encode information differently.
}
\label{fig:MIpatterns}
\end{figure}

\textbf{Note S4. Mutual Information Function}

The following equation details how Mutual Information was calculated between the RGB inputs to our system and the nine outputs observed from the microfluidic chip.

    \[
    MI(I; O) = \iint p(I, O) \log \left( \frac{p(I, O)}{p(I)p(O)} \right) \, dI \, dO,
    \]

$I$ is the input into the reservoir and can take one of the values $R$, $G$, or $B$. $O$ is the output of the reservoir and can be red, green, or blue from areas $1$, $2$, or $3$. These inputs and outputs to the reservoir have the MI calculated between them and are then displayed based on the quantization interval used. If a quantization interval of $2$ is used (Such as Figure 5(A)), then each cell in the heatmap is the product of multiple calculations comparing all input values and all output values quantized at two intervals. All of these cells make up the $3 \times 9$ heat map that we analyze for MI in the system.

\begin{figure}[H]\vspace{-1em}
\vspace{-1em}
\centering
\includegraphics[width=.5\textwidth]{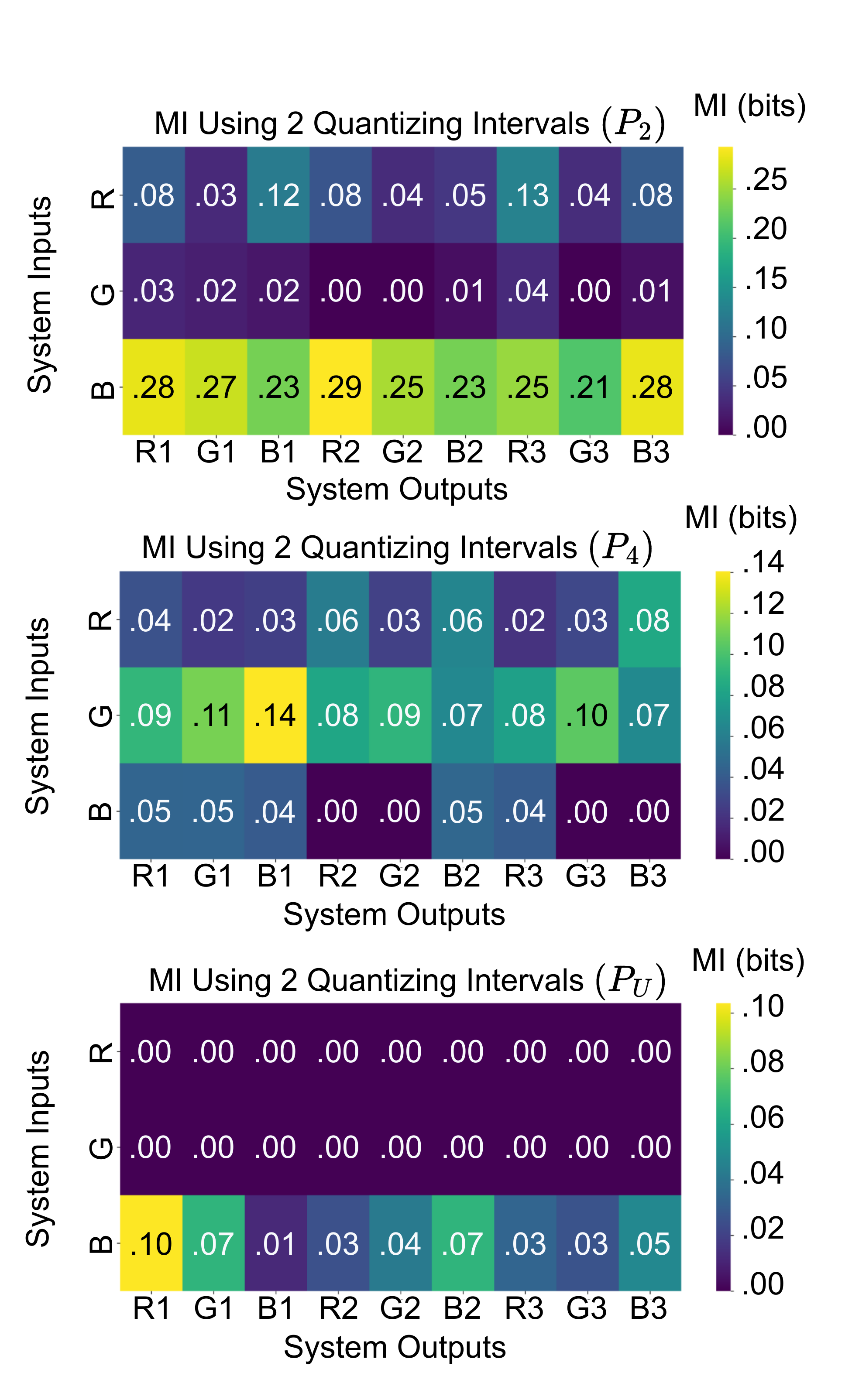}
\vspace{-1em}
\caption{
These MI plots are supporting the results in Figure 5(A) and are calculated per individual patterns rather than using all patterns. They show MI between the microfluidic chip's three inputs and nine quantized outputs, with the outputs represented as R, G, and B over three areas. The quantization interval for these plots is two. This is done to describe how specific patterns might behave in the system when quantized and where their information is located. 
}
\vspace{-1em}
\label{fig:2_quantizing_interval_MI}
\end{figure}

\textbf{Note S5. MI Analysis for Patterns $\bf{P_2}$, $\bf{P_4}$ \& $\bf{P_U}$ Using a Quantization Interval of $2$}

The MI analysis is evaluated for all patterns between the reservoir input and quantized reservoir output dye intensity signals flowing to and from the microfluidic reservoir. In this section, we focus specifically on patterns $\bf{P_2}$, $\bf{P_4}$, and $\bf{P_U}$ to observe how specific patterns behave in the system and which pairs of reservoir inputs and reservoir outputs carry more MI than others. MI is calculated between the three inputs to the microfluidic chip and nine outputs, with the outputs represented as R, G, and B in three areas. These details are provided in Figure~\ref{fig:2_quantizing_interval_MI}. We use quantized data instead of raw reservoir output to assess the impact of quantization on MI values. The heatmap for pattern $\bf{P_2}$ shows higher MI values between the blue input and all output areas, with all values above $0.20$ bits. This contrasts starkly with the green input, where the highest MI value among the output areas is only $0.04$ bits. This behavior results from blue dye being input to the system first, whereas in the second half of the pattern, red, green, and blue dyes are all present. When quantizing at two intervals, a significant amount of information is lost in the system. For pattern $\bf{P_4}$, higher MI values are associated with the green signal, the highest value appearing in the blue output of area $1$. The lowest MI values are associated with the blue input, which shows the MI is $0$ bits on four different outputs. When considering all variations in pattern $\bf{P_4}$, the green input appears to exhibit more fluctuation than the blue input, which may explain its stronger representation in the MI heatmap. Finally, for pattern $\bf{P_U}$ quantized at two intervals, there is a MI value of $0$ bits in the red and green inputs. The only MI appears between the blue input and all output areas. This outcome reflects the dominance of blue dye in the pattern, especially when low-resolution quantization is used.

\textbf{Note S6. MI Analysis for Patterns $\bf{P_2}$, $\bf{P_4}$ \& $\bf{P_U}$ Using a Quantization Interval of $5$} 

\begin{figure}[H]
\vspace{-1em}
\centering
\includegraphics[width=.5\textwidth]{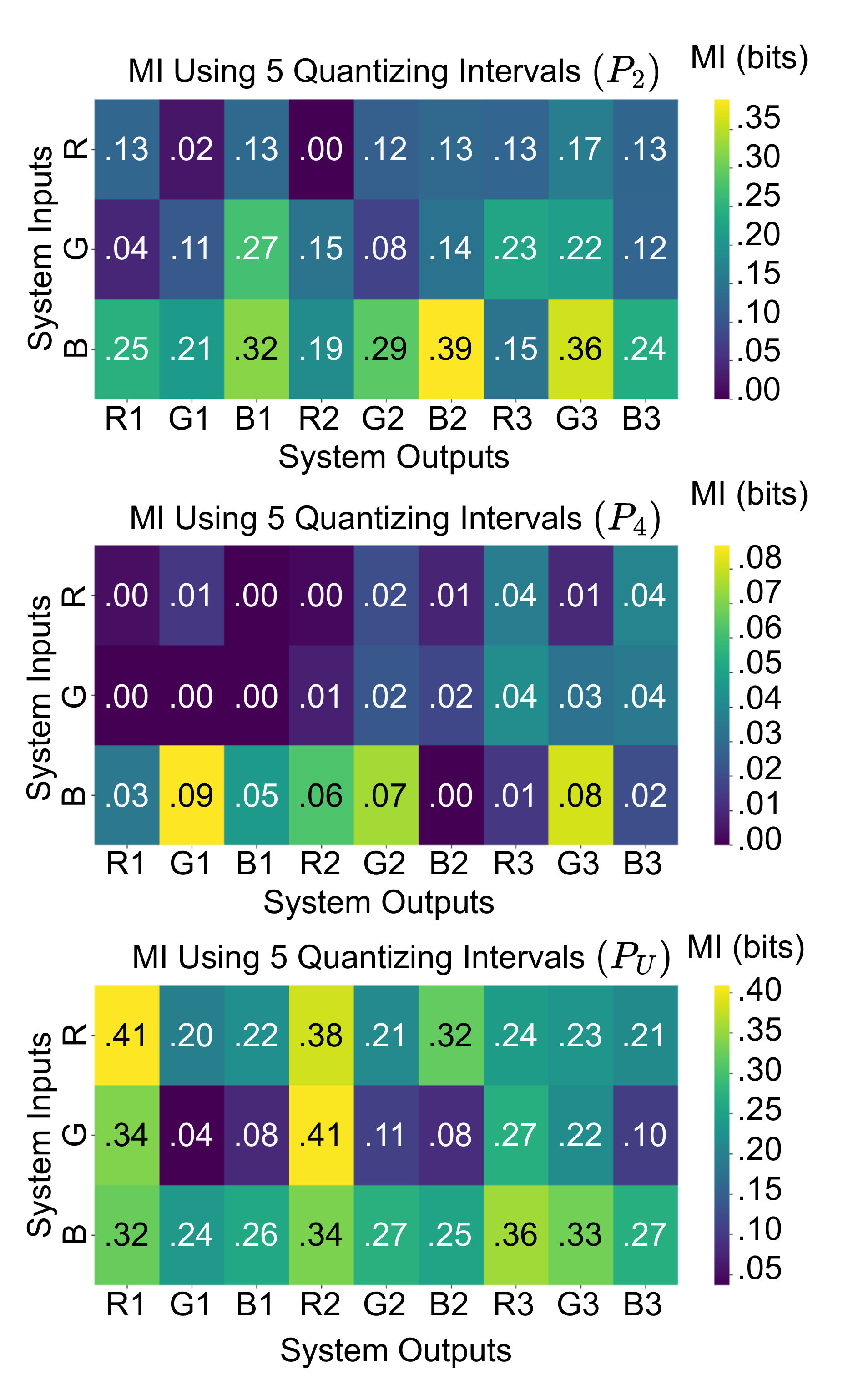}
\vspace{-1em}
\caption{
Similar to the $2$ quantization interval heatmaps, this set of heatmaps uses $5$ quantization intervals. These MI plots are supporting the results in Figure 5(B) and are calculated per individual patterns rather than using all patterns. They show MI between the microfluidic chip's three inputs and nine quantized outputs, with the outputs represented as R, G, and B over three areas. This is done to describe how specific patterns might behave in the system when quantized and where their information is located. 
}
\vspace{-1em}
\label{fig:5_quantizing_interval_MI}
\end{figure}

Similar to Note S5, this section examines the MI for patterns $\bf{P_2}$, $\bf{P_4}$ and $\bf{P_U}$ (displayed in Figure~\ref{fig:MIpatterns})using $5$ quantization intervals as shown in Figure~\ref{fig:5_quantizing_interval_MI}. Increasing the number of intervals from $2$ to $5$ provides more detailed MI heatmaps. For pattern $\bf{P_2}$, this increase is evident across the entire plot, with the blue input displaying an MI range of $0.15$ to $0.39$ bits in the top heatmap of Figure~\ref{fig:5_quantizing_interval_MI}. This contrasts with the narrower range observed at two quantization intervals, which spans only $.21$ to $.29$ bits in the top heatmap of Figure~\ref{fig:2_quantizing_interval_MI}. The highest MI value, $0.39$ bits, is found between the blue input and the blue output of area $2$. The green input also shows an increase in MI across all reservoir outputs. At $2$ quantization intervals, the highest MI value for the green input was $0.04$ bits, whereas with $5$ quantization intervals, it reaches $0.27$ bits in area $1$'s blue output. This increase results from the finer resolution of information captured by using more quantization intervals. The stronger influence of the green input on the blue output in area $1$ is due to green dye displacing blue dye during the second and third time steps of the pattern.

For pattern $\bf{P_4}$ shown in Figure~\ref{fig:MIpatterns}, we observe a major shift in MI values. At $2$ quantization intervals, the highest MI appears between the green input and all output areas. However, with $5$ quantization intervals, the blue input shows higher MI across all outputs, the peak value being $0.09$ bits between the blue input and area $1$'s green output. This aligns with the analysis in Figure 3(C-E), where constant signals contribute less information. When one dye is injected while the other is not in any pattern, the dye not being injected gets forced out, leading to information transfer in the system. Similar effects are seen in output area $2$ and $3$, where the MI values between the blue input and green outputs are $0.07$ bits and $0.08$ bits, respectively. Across the three output areas, MI is highest between the blue input and three green outputs, suggesting that this interaction is a key driver of information transfer in the system.

Finally, when analyzing pattern $\bf{P_U}$ displayed in Figure~\ref{fig:MIpatterns} at $5$ quantization intervals, MI is substantially higher than at $2$ intervals with an increase of $0.31$ bits when comparing the highest values from both intervals. Notable values include $0.41$ bits between the red input to the system and the red output of area $1$, and a similarly high value between the green input to the system and the red output of area $2$. The red input strongly correlates with area $1$ due to the larger channels that carry the red dye to that location, while the green input affects area $2$ due to its proximity to that output. The blue input again exhibits high MI across all output areas, reflecting its greater presence in the pattern, as it was injected four times, compared to only twice for the red and green inputs. These heatmaps clearly demonstrate the sharp increase in information captured when the number of quantization intervals is increased from $2$ to $5$. This additional resolution enables a more nuanced interpretation of system dynamics.

\begin{figure}[H]
\vspace{-1em}
\centering
\includegraphics[width=.5\textwidth]{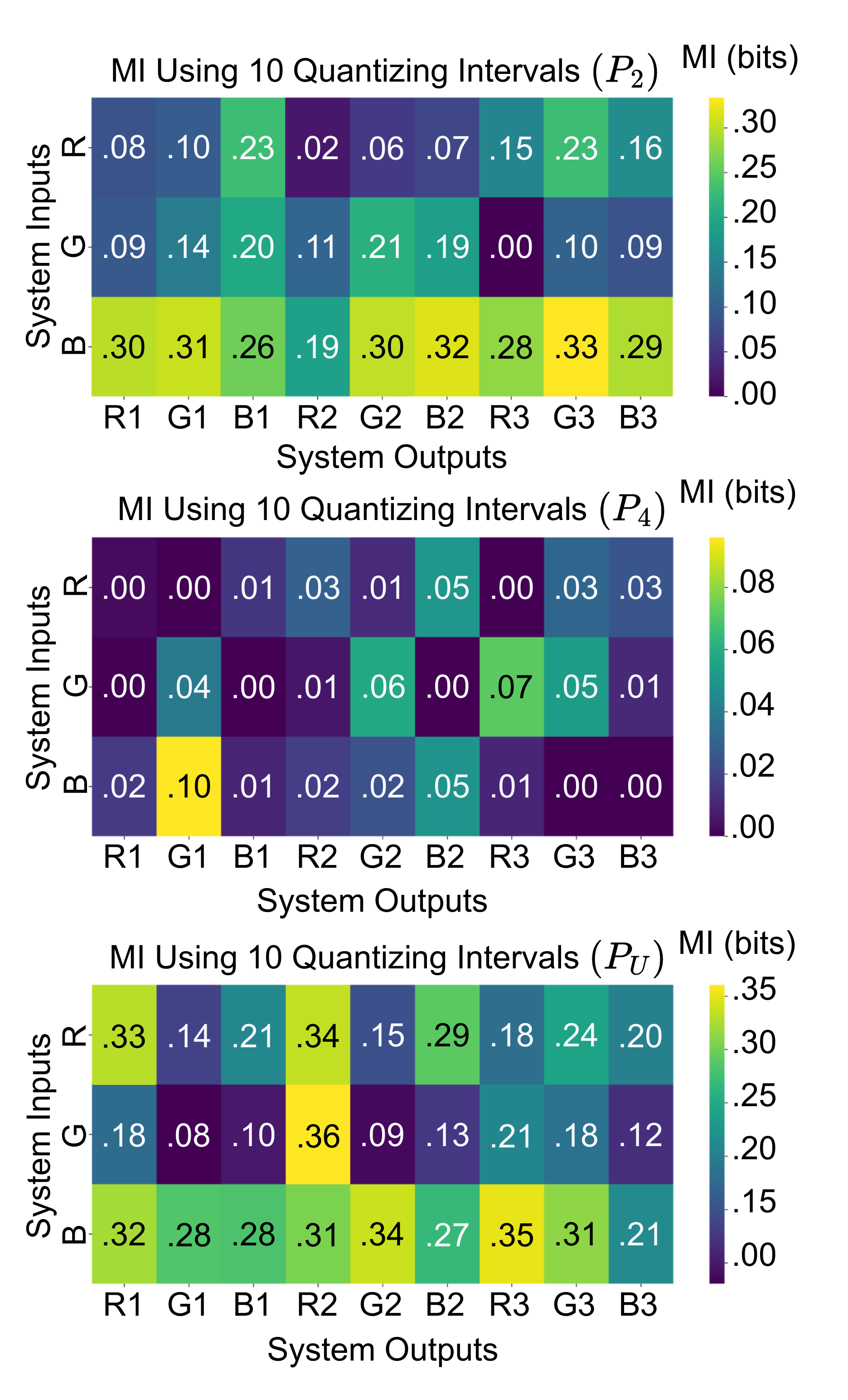}
\vspace{-1em}
\caption{
Similar to Figure S4 and S5, this set of heatmaps uses $10$ quantization intervals. These MI plots are supporting the results in Figure 5(C) and are calculated per individual patterns rather than using all patterns. They show MI between the microfluidic chip's three inputs and nine quantized outputs, with the outputs represented as R, G, and B over three areas. This is done to describe how specific patterns might behave in the system when quantized and where their information is located. 
}
\vspace{-1em}
\label{fig:10_quantizing_interval_MI}
\end{figure}

\textbf{Note S7. MI Analysis for Patterns $\bf{P_U}$, $\bf{P_4}$ \& $\bf{P_U}$ Using a Quantization Interval of $10$}

Similar to Note's S5 and S6, this Note examines MI in the microfluidic reservoir using ten quantization intervals. These heatmaps follow trends similar to those seen when using a quantizing interval of $5$: the blue input in pattern $\bf{P_2}$ shows high MI values across all output areas in the top heatmap of Figure~\ref{fig:10_quantizing_interval_MI}; the highest MI value in pattern $\bf{P_4}$ occurs between the blue input to the system and the green output of area $1$ in the middle heatmap of Figure~\ref{fig:10_quantizing_interval_MI}; and finally, in pattern $\bf{P_U}$, the blue input again shows high MI across all outputs, with additional high MI between the red input and the red output of area $1$ in the bottom heatmap of Figure~\ref{fig:10_quantizing_interval_MI}, as well as the green input and red output of area $2$ on the same heatmap. Looking at the specific MI values in each heatmap, pattern $\bf{P_2}$ shows a peak MI of $0.33$ bits between the blue input and the green output of area $3$. The lowest value is $0$ bits, between the green input and the red output of area $3$. In pattern $\bf{P_4}$, most MI is associated with the blue input and the green output of area $1$ with a value of $0.10$ bits, as well as the green input and the red output from area $3$ with a value of $0.07$ bits. The heatmap for pattern $\bf{P_U}$ shows a broad spread of MI values across all inputs and outputs, with the blue input maintaining consistent MI across all outputs. The green input reaches its highest MI value between it and red output of area $2$. These results are consistent with the findings in Figure 3(C-E): patterns with varying and alternating signals retain more information through the reservoir, while constant input patterns result in minimal information transfer.

\begin{figure}[H]
\vspace{-1em}
\centering
\includegraphics[width=1\textwidth]{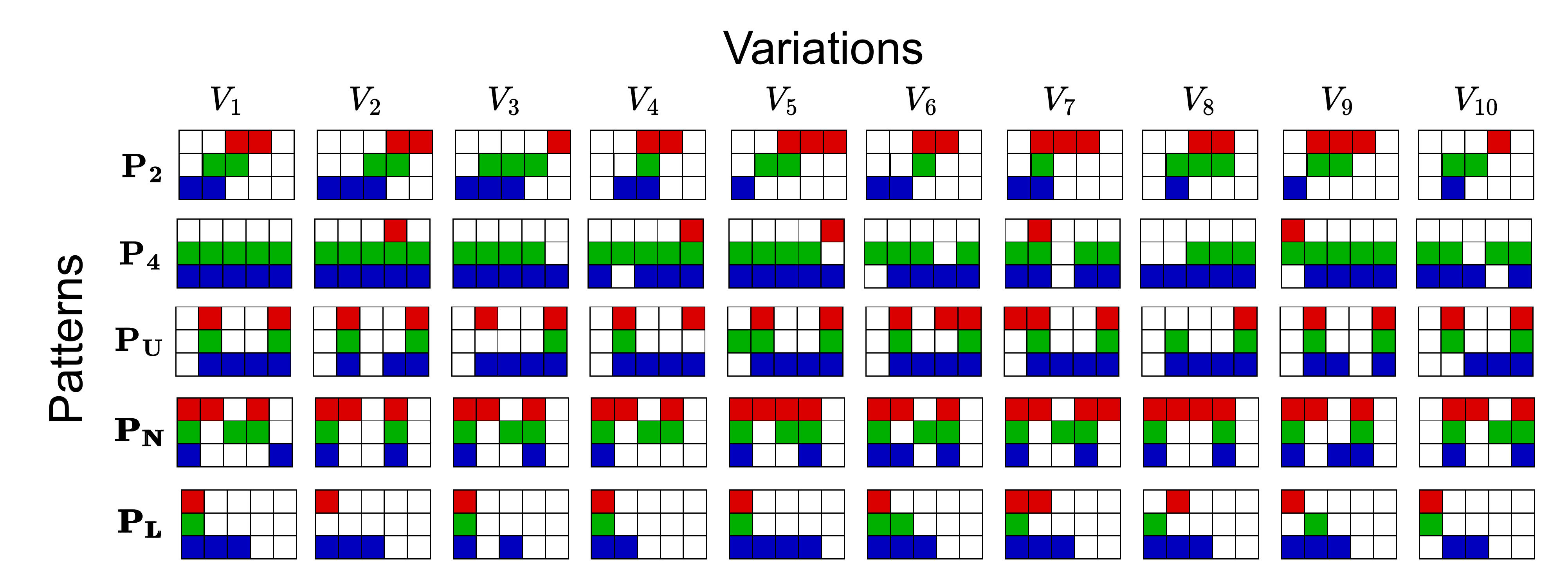}\vspace{-2em}
\caption{
Patterns that are being compared in Figure\ref{fig:N_U_Similarity}\&\ref{fig:N_L_Similarity}\&\ref{fig:2_4_Similarity}
}
\vspace{-1em}
\label{fig:Pattern_Variation_Patterns}
\end{figure}

\textbf{Note S8. Similarity Comparison of Patterns $\bf{P_2}$, $\bf{P_4}$, $\bf{P_U}$, $\bf{P_N}$, and $\bf{P_L}$}

\begin{figure}[H]
\vspace{-1em}
\centering
\includegraphics[width=0.7\textwidth]{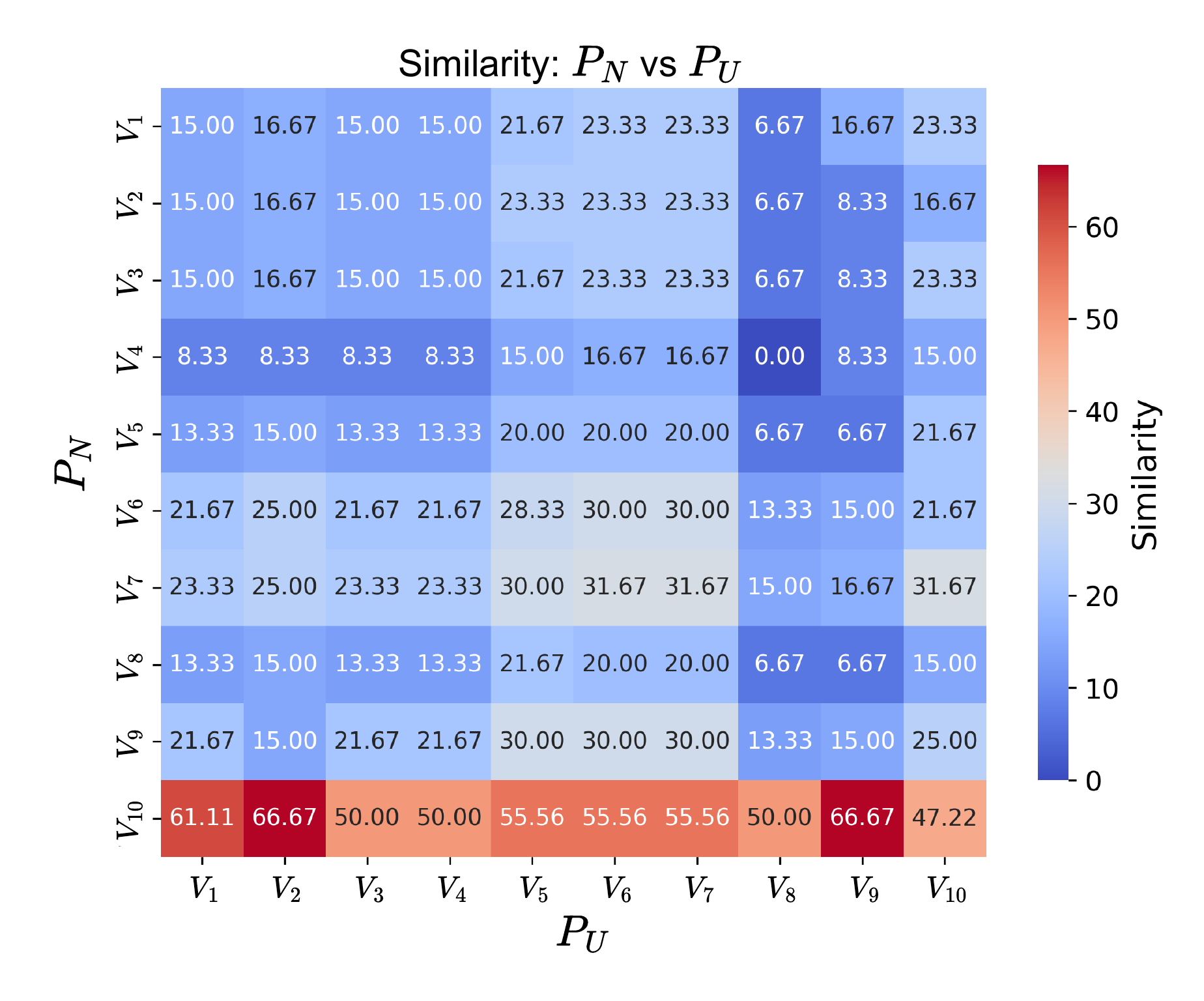}.
\vspace{-1em}
\caption{
A heatmap showcasing the similarity between patterns $\bf{P_N}$ and $\bf{P_N}$. While most variations of pattern $\bf{P_N}$ are not similar to pattern $\bf{P_U}$, variant $P_{N,V_{10}}$ has a high similarity to all of pattern $\bf{P_U}$, reaching $67\%$ similarity. This shows that variant $P_{N,V_{10}}$ is not a good match for the dataset and can result in misclassifications.
}
\vspace{-1em}
\label{fig:N_U_Similarity}
\end{figure}

\begin{figure}[H]
\vspace{-1em}
\centering
\includegraphics[width=0.7\textwidth]{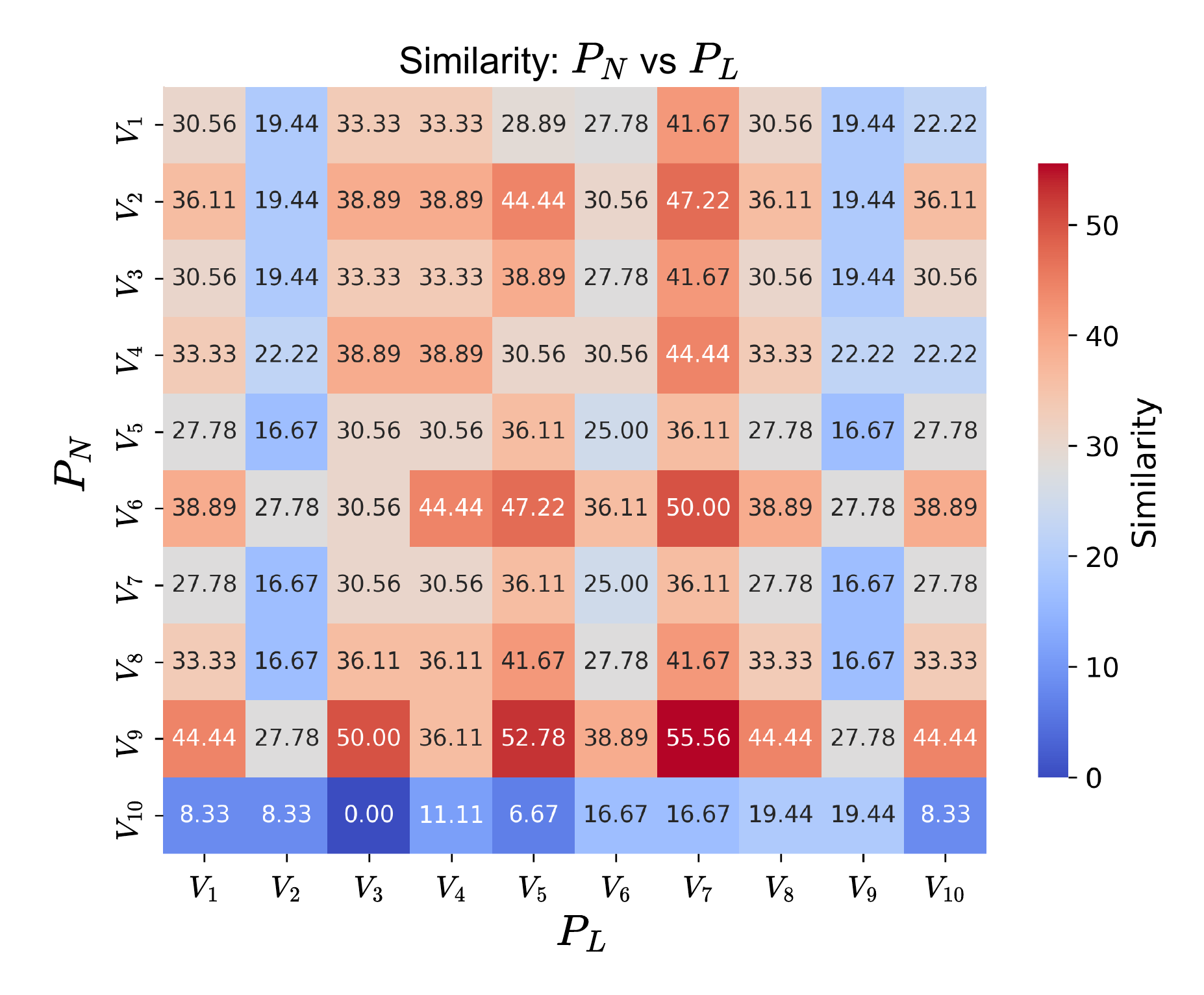}
\vspace{-1em}
\caption{
A heatmap showcasing the similarity between patterns $\bf{P_N}$ and $\bf{P_L}$. 
In this comparison analysis, most variations of pattern $\bf{P_N}$ are at least $30\%$ similar to the variations of pattern $\bf{P_L}$. Variant $P_{N,V_{10}}$, however, is not similar to pattern $\bf{P_L}$. This suggests that it is not a good representation of pattern $\bf{P_N}$ since it does not follow the trends of the other variations, and should be treated as an outlier in our dataset.
}
\vspace{-1em}
\label{fig:N_L_Similarity}
\end{figure}

\begin{figure}[H]
\vspace{-1em}
\centering
\includegraphics[width=0.7\textwidth]{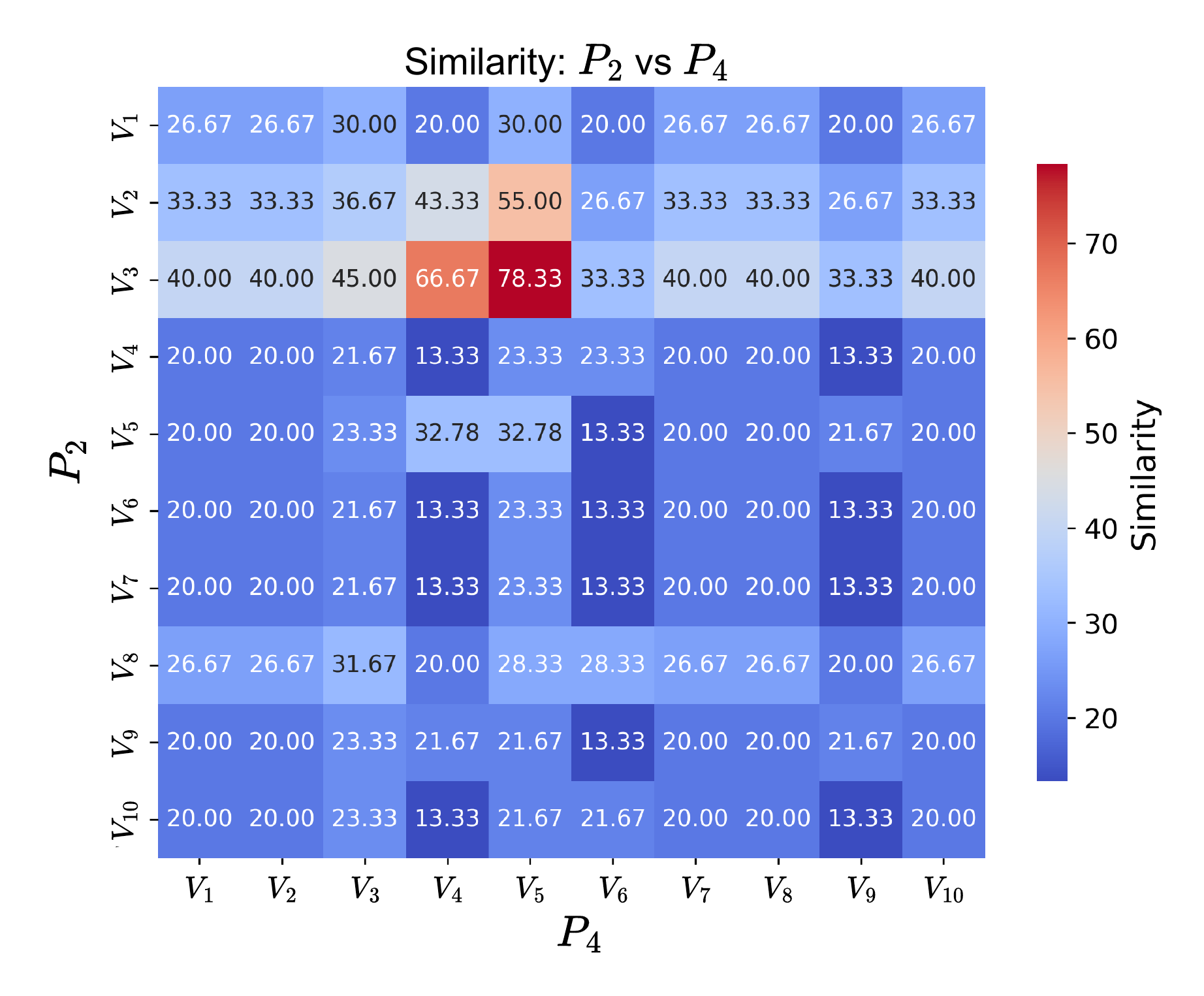}
\vspace{-1em}
\caption{
A heatmap showcasing the similarity between patterns $\bf{P_2}$ and $\bf{P_4}$. A majority of the variations for patterns $\bf{P_2}$ and $\bf{P_4}$ have a $20\%$ match, however, variants $P_{2,V_3}$ and $P_{4,V_5}$ share a $78.33\%$ percent match, extremely high for two patterns that are supposed to be different. This similarity could lead to misclassifications in our system.
}
\vspace{-1em}
\label{fig:2_4_Similarity}
\end{figure}

Comparing patterns is essential in investigating how similarity an internal consistency between variations affects accuracy. Figure~\ref{fig:Pattern_Variation_Patterns} displays the patterns under analysis, where patterns $\bf{P_U}$, $\bf{P_N}$, and $\bf{P_L}$ appear in the top row and are included to evaluate how similar the letter-based patterns are. Patterns $\bf{P_2}$ and $\bf{P_4}$, which are not character-based, are also included to assess how non-letter patterns behave under the same similarity analysis. Finally, variation $P_{N, V_{10}}$ is included due to repeated misclassifications in trained the models.
Figure~\ref{fig:N_U_Similarity} presents the similarity across all variations for patterns $\bf{P_N}$ and $\bf{P_U}$. Variants $\bf{P_{N,V_{1}}}$ to $\bf{P_{N,V_{9}}}$ show less than $32\%$ similarity to 
pattern $\bf{P_U}$, while variant $P_{N,V_{10}}$ reaches a $66.67\%$ match to pattern $\bf{P_U}$. This trend is seen in Figure~\ref{fig:N_L_Similarity}, where variants $\bf{P_{N,V_{1}}}$ to $\bf{P_{N,V_{9}}}$ show high similarity to pattern $\bf{P_L}$ reaching up to $55.56\%$, while variant $P_{N,V_{10}}$, which deviates from other variations, shows a $19\%$ match at its highest. This behavior tells us that variant $P_{N,V_{10}}$ is not a good representation of pattern $\bf{P_N}$ due to not following trends shown by other variations.
Further analysis of this discrepancy, in Figure~\ref{fig:Pattern_Variation_Patterns}, we observe that variant $P_{N,V_{10}}$ shifts to the right and overlaps with pattern $\bf{P_U}$. This is the only example of shifting in the dataset, making variant $P_{N,V_{10}}$ an outlier for our system. With our currently limited dataset and resolution, this variation causes classification errors.
There is a separate trend when analyzing the similarity between patterns $\bf{P_2}$ and $\bf{P_4}$ in Figure \ref{fig:2_4_Similarity}. With these two patterns, most variations fall between $20\%$ and $40\%$ similarity. However, variants $P_{2,V_3}$ and $P_{4,V_5}$ have a $78.33\%$ match. Unlike the issue with variant $P_{N,V_{10}}$, this similarity spike occurs despite variants $P_{4,V_5}$ and $P_{2,V_3}$ accurately representing their respective patterns. This highlights the challenge of working with low-resolution ($3 \times 5$) data; small shifts in pixel variation can make one pattern resemble another. A potential solution is to increase the image resolution, which may help preserve distinguishing features across variations.

\textbf{Note S9. Mean Absolute Difference Analysis of Patterns $\bf{P_4}$, $\bf{P_N}$, and $\bf{P_L}$} 

The purpose of the mean absolute difference (MAD) comparison is to evaluate how similar the reservoir outputs are for different patterns and their variations after passing through the microfluidic reservoir. In Figure~\ref{fig:N_L_4_Mean-Absolute_Difference}, patterns $\bf{P_N}$ and $\bf{P_L}$ are compared in the first heatmap. Notably, variant $P_{L,V_7}$ shows a significantly smaller difference with pattern $\bf{P_N}$ than the other variations, suggesting that the system may confuse it with pattern $\bf{P_N}$.
In contrast, the second heatmap in Figure~\ref{fig:N_L_4_Mean-Absolute_Difference} compares patterns $\bf{P_N}$ and $\bf{P_4}$, where the differences between the two are more distinct. This is in part because pattern $\bf{P_N}$ contains some red dye, while pattern $\bf{P_4}$ only includes green and blue dye, a clear differentiating factor. The closest match between these two patterns occurs between variants $P_{4,V_8}$ and $P_{N,V_{10}}$, with a MAD of $22.95\%$. This close match results from variations that cause both patterns to overlap more in structure, making their reservoir outputs appear similar. As discussed previously, variant $P_{N,V_{10}}$ is a poor representation of the dataset and is considered an outlier. Most other variations of pattern $\bf{P_N}$ show MAD values above $40\%$, reinforcing this point. These comparisons highlight how variations within a pattern can affect classification accuracy, as certain variations more closely resemble those of other patterns after being sent through the reservoir.

\begin{figure}[H]
\vspace{-1em}
\centering
\includegraphics[width=.8\textwidth]{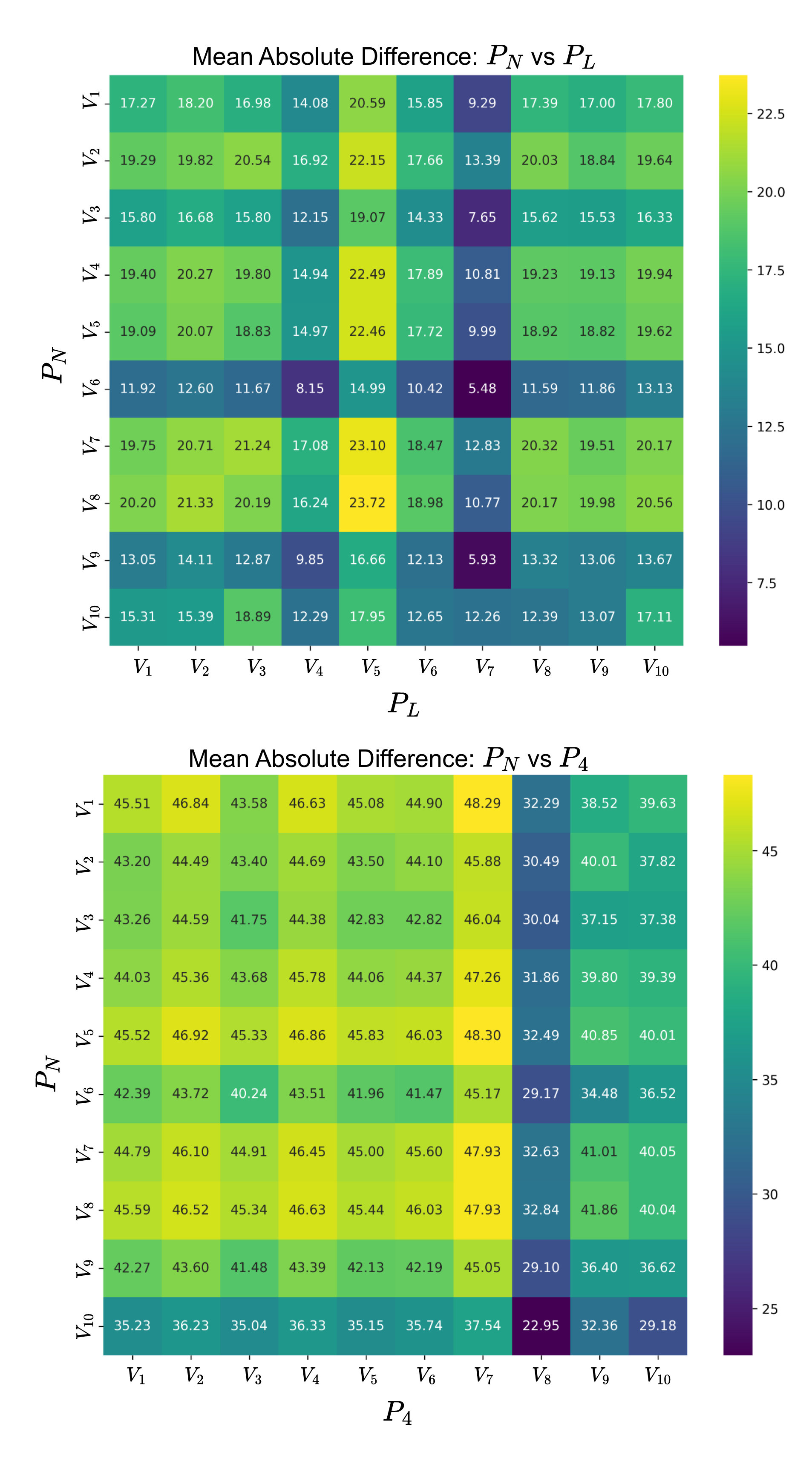}
\label{fig:N_L_Mean-Absolute_Difference}
\end{figure}
\pagebreak
\begin{figure}[t!]
\caption{(Previous page) This image shows both heatmaps comparing pattern $\bf{P_N}$ with patterns $\bf{P_L}$ and $\bf{P_4}$ with respect to the mean absolute difference. This analysis is conducted on reservoir outputs before they have been quantized. For patterns $\bf{P_N}$ and $\bf{P_L}$, there is little difference between variants $P_{L,V_7}$ and $P_{N,V_{6,9}}$ This led to confusion when classifying these pattern variations. For patterns $\bf{P_N}$ and $\bf{P_4}$, there is minimal similarity with the closest mean absolute difference between patterns being $22.95$. This means that these patterns will have a low probability of being mis-categorized.
}
\vspace{-1em}
\label{fig:N_L_4_Mean-Absolute_Difference}
\end{figure}

\begin{figure}[H]
\centering
\includegraphics[width=1\textwidth,angle=-90]{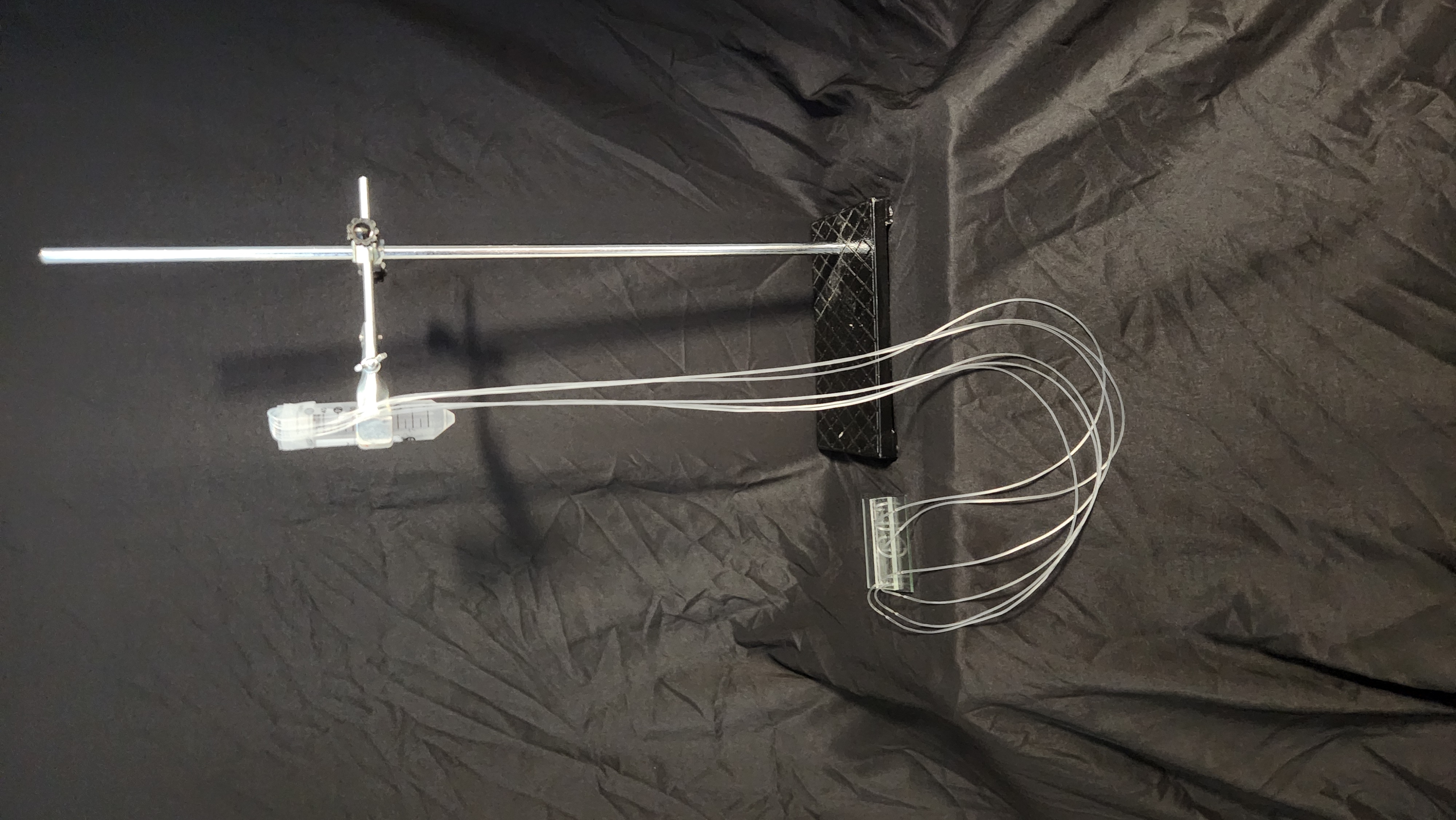}
\caption{
A hydrostatic setup for getting air out of the \emph{polydimethylsiloxane} (PDMS) microfluidic chip.
}
\vspace{-1em}
\label{fig:Hydrostatic}
\end{figure}

\textbf{Note S10. Hydrostatic Setup} 
The hydrostatic setup is used to remove air from the microfluidic chip prior to testing. Air pockets can block the channels and prevent liquid from flowing properly, so the chip must first be filled with water to enable free movement of dye during experiments. Six tubes are connected to all the inlet and outlet ports of the chip, each linked to a water-filled container. This container is positioned above the microfluidic chip, allowing gravity to drive water into the channels and push air out through the \emph{polydimethylsiloxane} (PDMS) material.

\bibliography{bibliography}